\documentclass{article}

\usepackage{arxiv}
\pdfoutput=1                
\usepackage[utf8]{inputenc} 
\usepackage[T1]{fontenc}    
\usepackage{hyperref}       
\usepackage{url}            
\usepackage{booktabs}       
\usepackage{amsfonts}       
\usepackage{nicefrac}       
\usepackage{microtype}      
\usepackage{lipsum}
\usepackage{graphicx}
\usepackage{placeins}
\usepackage[table,xcdraw]{xcolor}
\usepackage{multirow}

\title{Prediction Model for Semitransparent Watercolor Pigment Mixtures Using Deep Learning with a Dataset of Transmittance and Reflectance}

\author{
    Mei-Yun~Chen\\
  Graduate Institute of Networking \& Multimedia\\
  National Taiwan University\\
  \texttt{meiyun@cmlab.csie.ntu.edu.tw} \\
  \And
   Ya-Bo~Huang\\
   Department of Computer Science \& Information Engineering\\
   National Taiwan University\\
   \texttt{pixiyon@cmlab.csie.ntu.edu.tw}
   \And
   Sheng-Ping~Chang \\
   NTU IoX Center \\
   National Taiwan University \\
   \texttt{shanechang@ntu.edu.tw}\\
   \And
   Ming~Ouhyoung \\
   Graduate Institute of Networking \& Multimedia\\
   National Taiwan University\\
  \texttt{ming@csie.ntu.edu.tw}\\
}

\begin{document}
\maketitle

\begin{abstract}

Learning color mixing is difficult for novice painters. In order to support novice painters in learning color mixing, we propose a prediction model for semitransparent pigment mixtures and use its prediction results to create a Smart Palette system. Such a system is constructed by first building a watercolor dataset with two types of color mixing data, indicated by transmittance and reflectance: incrementation of the same primary pigment and a mixture of two different pigments. Next, we apply the collected data to a deep neural network to train a model for predicting the results of semitransparent pigment mixtures. Finally, we constructed a Smart Palette that provides easily-followable instructions on mixing a target color with two primary pigments in real life: when users pick a pixel, an RGB color, from an image, the system returns its mixing recipe which indicates the two primary pigments being used and their quantities. 

When evaluating the pigment mixtures produced by the aforementioned model against ground truth, 83$\%$ of the test set registered a color distance of $\Delta E^{*}_{ab}$ $<$ 5; $\Delta E^{*}_{ab}$, above 5 is where average observers start determining that the colors in comparison as two different colors. In addition, in order to examine the effectiveness of the Smart Palette system, we design a user evaluation which untrained users perform pigment mixing with three methods: by intuition, based on Itten's color wheel, and with the Smart Palette and the results are then compiled as three color distance, $\Delta E^{*}_{ab}$ values. After that, the color distance of the three methods are examined by a t-test to prove whether the color differences were significant. Combining the results of color distance and the t-values of the t-test, it can demonstrate that the mixing results produced by using the Smart Palette is obviously closer to a target color than that of the others. Base on these evaluations, our system, the Smart Palette demonstrates that it can effectively help users to learn and perform better at color mixing than that of the traditional method.   

\end{abstract}

\keywords{Color mixing \and Itten's color wheel \and Color matching \and Reflectance \and Transmittance \and Spectrometer \and Deep neural network}

\section{Introduction}

Color mixing is difficult for watercolor painting novices and the most common learning method is through Itten's color wheel \cite{itten1970elements}. It provides the very basic concept on what happens when pigments combined with 1:1 ratio. It is, however, a tedious process with a clear obstacle: Itten's wheel display too few cases of color mixing. Therefore, using Itten's color wheel for acquiring color mixing skill is difficult. 

Many studies on color mixing are based on a method with three procedures. First, the pigment absorption and scattering \cite{kubelka1931article, kubelka1948new} can be determined through the Kubelka-Munk (KM) theory; then by employing the Duncan formula \cite{duncan1940colour} to calculate the absorption and scattering of pigments, the mixture result of absorption and scattering can be obtained. Finally, the previous result is converted to reflectance using KM theory. For example, Haase and Meyer \cite{haase1992modeling} use this method to compute the colors of arbitrary pigment mixtures for generating realistic images in the computer graphics domain. Baxter et al. \cite{baxter2004impasto} improve color accuracy in their painting system using this method. Furthermore, Aharoni-Mack et al. \cite{aharoni2017pigment} calculate a pigment palette for recoloring watercolor paintings using this method.

In the previously mentioned studies which employ the Duncan formula \cite{duncan1940colour}, calculation for color mixing uses pigment ratios as the parameter. In real life, however, because some media have high transparency, which means adding more quantities of pigments, despite the mixture ratio remains, will actually cause color change. It is especially the case for watercolor, because it a media of high transparency and the relation of pigment and substrate must be considered. Therefore, in this study, we propose a model, which additionally takes transmittance of pigments into consideration, for predicting the mixtures of physical watercolor pigments and apply the prediction results to create a Smart Palette for helping novices to learn pigment mixing easily.

Our proposed model is built with the following steps: First, we construct a watercolor dataset measured by a spectrometer; data set contains reflectance and transmittance of 13 primary pigments (sampling 12 different quantities each), and the reflectance of 780 mixture results from two arbitrary primary pigments. Then, we use a deep neural network (DNN) to train our prediction model, which predicts the mixture of two primary pigments. Next, we apply the prediction results to construct a look-up table. Lastly, we design a system, a Smart Palette, which has two functions: color matching and color mixing.

This study has the following contributions:

\begin{enumerate}
\item  It constructs a watercolor dataset named NTU watercolor pigments spectral measurement dataset, hereinafter referred to as NTU WPSM dataset. We employ a spectrometer, OTO SD1220, using a light source box and K1 light source to measure the transmittance and reflectance of primary pigment samples, as well as the reflectance of the pigment mixture samples; then, the measured data are labeled. NTU WPSM dataset has two types of data: 

\begin{description} 
\item a. The first type of dataset is collected from adding the same primary pigments to itself, namely increasing the thickness/quantities of a primary pigment. Hereinafter, Data Type I. (I for incrementation)
\item b. The second type of dataset is collected by mixing two different primary pigments with three mixing ratio (1;1, 1:2, and 2:1) in different quantities. (In other words, mixing 0.01 mL of Pigment A with 0.01 mL of Pigment B is counted as a different sample from mixing 0.02 mL of Pigment A with 0.02 mL of Pigment B.) Hereinafter, Data Type M. (M for mixing)
\end{description}
\item It constructs a prediction model of semitransparent watercolor pigment mixtures, hereinafter referred to as SWPM prediction model. We train SWPM prediction model using a DNN with Google tensorflow for implementation. SWPM prediction model takes 2 primary pigments, with specified quantity, and predicts the mixture result.  

\item It constructs a Smart Palette. We provide a software tool of color matching for amateur painters where the input is a desired color, and the output is a recipe containing information on the two primary pigments and their individual quantities that could be used to recreate the input desire color; this is important for an amateur painter because when one makes a mistake in mixing colors, adding more pigments generally results in the output becoming an undesired dark color. In addition, the tool has a second function: color mixing. We utilize the prediction results of SWPM prediction model to allow users to simulate results of color mixing in the system.
\end{enumerate} 
\section{Related work}

\subsection{Color mixing theory of Itten's color wheel}

\begin{figure}[htbp]
\centering
  \includegraphics[width = 1.0\textwidth]{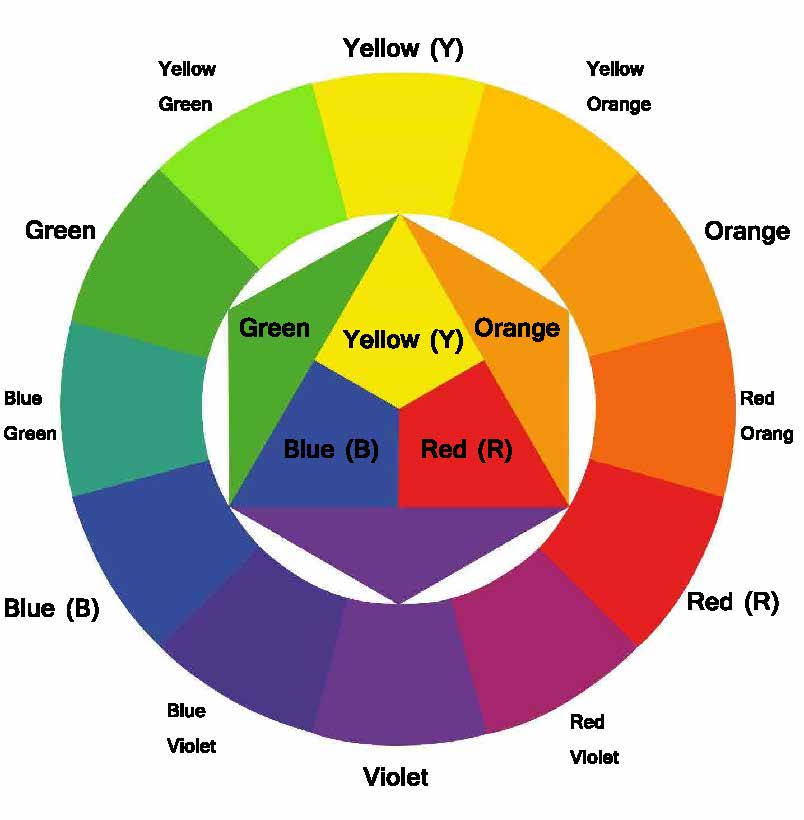}
\caption{Itten 12-part color wheel. Mixing the primary colors blue, red, yellow with 1:1 ratio to receive the secondary colors: orange, green, and violet. Then the secondary colors are mixed with the primary colors with, also with 1:1 ratio, to form the tertiary colors: yellow-orange, red-orange, red-violet, blue-violet, blue-green, and yellow-green.}
\label{fig:Itten}       
\end{figure}

\begin{figure}[htbp]
\centering
  \includegraphics[width = 1.0\textwidth]{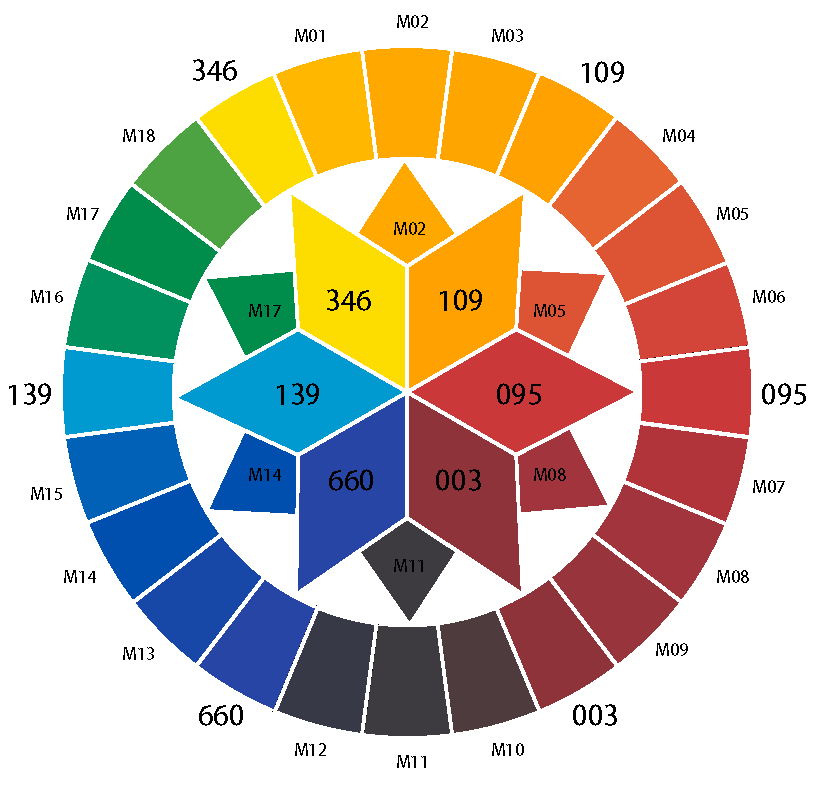}
\caption{The 24-part color wheel we modeled with Winsor and Newton pigments. It is created following the same principle that of Itten's, where M02, M05, M8, M11, M14, and M17 are the secondary colors; M01, M03, M04, M06, M07, M09, M10, M12, M13, M15, and M16 are the tertiary colors.}
\label{fig:ColorWheel_Mix}       
\end{figure}

In term of studying color mixing, the most common way nowadays is still based on the color theory developed by Johannes Itten\cite{itten1970elements} in the 1970s. The famous color wheel (Fig. \ref{fig:Itten}) shows the relationships between 2 colors at a time, typically starts the three primary colors blue, red, and yellow at the center and expand outward to secondary and the tertiary color layers. It provides the very basic concept on what happens when two pigments combined with 1:1 ratio to learners.  
This theory, although being the most common learning method, has one major limitation itself and thus also limits the learning effectiveness: the mixture examples can be efficiently displayed by the wheel is just too few. As example, we develop one with Winsor and Newton pigments replacing the three primary colors with 6 primary-hue colors, as shown in Fig. \ref{fig:ColorWheel_Mix}. This attempt expands the quantity of pigment example for learning from the original 12 to 24 but the problem still persists. Because: a. 24 cases is still too few references for virtually limitless combination possibilities of pigment mixtures; and b. colors shown in one wheel is only accurate for showing the intermediate point of two adjacent colors, any combinations that skip, or on the opposite end of, the set arrangement order requires another new wheel to display accurate results.

%

%
  
\subsection{Color mixing studies in computer graphics}

When we discuss putting two or more colors together to produce a new color, there is actually a clear distinction between color compositing and color mixing: color compositing is the layer superimposition of pigments on a substrate through multiple applications, whereas color mixing is the pigment particles being mixed prior to being applied on a substrate. Both subjects are discussed by many different studies.

Studies on color compositing, for example, are discussed by Curtis et al. \cite{curtis1997computer} using KM theory and Kubelka's method \cite{kubelka1954new} to combine layers of watercolor pigments and a backing paper. Gossett and Chen \cite{gossett2004paint} employ RYB compositing to create a mixture, which is more intuitive than RGB compositing, to effectively visualize overlapping areas using an RYB cube and trilinear interpolation. Lu et al. \cite{lu2014realpigment} propose two prediction models for compositing pigment layers: radial basis function interpolation and optimized KM compositing; the result of the optimized KM compositing was deemed to behave as more paint-like by the team.

Studies on color mixing, including those in the computer graphics area, use KM theory to obtain the two constants which are absorption and scattering, $K$ and $S$, or one constant, $\frac{K}{S}$, by calculating measured reflectance \cite{kubelka1931article, kubelka1948new}; subsequently, they use the Duncan formula to compute color mixing \cite{duncan1940colour}. The results obtained via this method is called the two-constant/single-constant KM theories \cite{yang2010kubelka}.

The single-constant KM theory is used to predict an opaque pigment mixing \cite{davidson1966color} and the two-constant KM theory is effective for mixing nonopaque pigments. For example, Haase and Meyer \cite{haase1992modeling} use single-constant KM theory to mix pigments for realistic image synthesis. Mohammadi and Berns \cite{mohammadi2004verification} use single-constant and two-constant KM theory to predict color mixing of acrylic paint. Baxter et al. \cite{baxter2004impasto} calculate absorption and scattering using single-constant KM theory for predicting the mixture of the pigments in their painting system. Okumura \cite{okumura2005developing} constructs an artist paint database using two-constant KM theory. Aharoni-Mack et al. \cite{aharoni2017pigment} recolor watercolor paintings by using two-constant KM theory. In addition, Minato \cite{Minato1969reference} proposes a method, which is based on two-constant KM theory, to determine scattering with the assistance of two reference color samples. After that, Tominaga et al. \cite{tominaga2016modeling} apply Minato's method and combine the constants and the thickness, $X$, which means substituting $KX$ and $SX$ for $K$ and $S$, to estimate the reflectance of diluted watercolor pigments under various concentrations, and the mixing of diluted watercolor at the 50$\%$ water ratio.

Some studies have achieved color mixing by other methods as well. Xu et al. \cite{xu2007generic} use a neural network with three kernel functions, of which the first function is based on KM theory, to predict pigment mixing. The input components of the neural network are the concentration, absorption, scattering, transmittance, and reflectance of the two pigments to be mixed; except for the concentration, each component is denoted by three wavelengths: pure red, green, and blue. Xu et al. apply the reflectance of a measured pigment to estimate a term, $\frac{K}{S}$, using single-constant KM theory and assuming $S$ to be 1; thus, $K$ is known; moreover, the transmittance of the input pigments is calculated using KM theory. Tober \cite{tober2015colorigins} employs a mathematical model, the weighted geometric mean, to obtain color mixing results by calculating the reflectance of the primary pigments.

Our study involves color mixing. More specifically, we propose a prediction model to provide assistance to watercolor painter in color mixing.  In our study, we not only measure the reflectance of the watercolor pigments but also their transmittance. After constructing a dataset, we use a DNN to obtain a prediction model of semitransparent pigment mixtures. 
\section{Method}

\subsection{Dataset of watercolor pigments}
To construct SWPM prediction model ,we selected 13 primary pigments that are commonly favored by watercolorists to be the bases for our dataset. In this section, we first explain the method used in creating and measuring our samples from the selected 13 pigments (see Sect. \ref{MakeSamples}). Then, the labeling of the measured data for training prediction model is discussed (Sect. \ref{labelData}). 

\subsubsection{Making and measuring the samples of watercolor}
\label{MakeSamples}
First and foremost, the premise of SWPM prediction model is as follows: one can select 2 pigments from the 13 selected primary pigments as mixing ingredients and SWPM prediction model predicts how such watercolor mixture will result when applied onto a piece of white paper. 

Two features that are important and differentiate watercolor pigments from other media. First, watercolor pigment has highly transparent attribute and each pigment has different transparency. In other words, with watercolors, thicker pigment layers have greater opacity than thinner layers of the same pigment. Secondly, coating pigments in thin layers is crucial in making watercolor creations. Considering such features, we decide to add transmittance on top of reflectance in sample measurement and data denotation for each pigment. 

The construction for SWPM prediction model requires 7 components in total, 6 for the input layer; 1 for the output layer. Here, in a mixing process, we define the first selected pigment as $P_A$ and the second as $P_B$:
\begin{itemize}
\item Input layer:
\end{itemize}

\begin{enumerate}
\item Transmittance of a $P_A$ (primary pigment)
\item Reflectance of a $P_A$ (primary pigment)
\item Transmittance of a $P_B$ (primary pigment)
\item Reflectance of a $P_B$ (primary pigment)
\item Reflectance of substrate (white paper, Canson Ca grain)
\item Quantities of a $P_A$ and $P_B$ (primary pigment quantities)
\end{enumerate}

\begin{itemize}
\item Output layer:
\end{itemize}
\begin{enumerate}
\item Reflectance of the mixture result $P_A + P_B$
\end{enumerate}

\begin{figure}[htbp]
\centering
  \includegraphics[width = 1.0\textwidth]{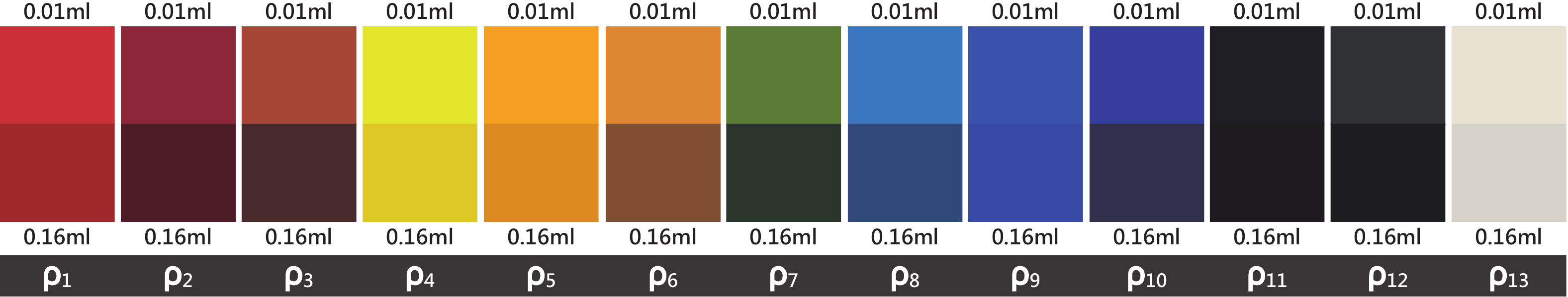}
\caption{Thirteen primary pigments. To correctly display their colors, we use the measured results. The top row shows the pigments in layers comprising 0.01 mL, and the bottom row shows the pigments in layers comprising 0.16 mL; when these two displays of one pigment have a larger color difference, it means that the pigment has high transparency, such as burnt sienna $\rho_3$, raw sienna $\rho_6$, sap green $\rho_7$, and cerulean blue $\rho_8$.}
\label{fig:13pigments}       
\end{figure}

\begin{figure}[htbp]
\centering
  \includegraphics[width = 1.0\textwidth]{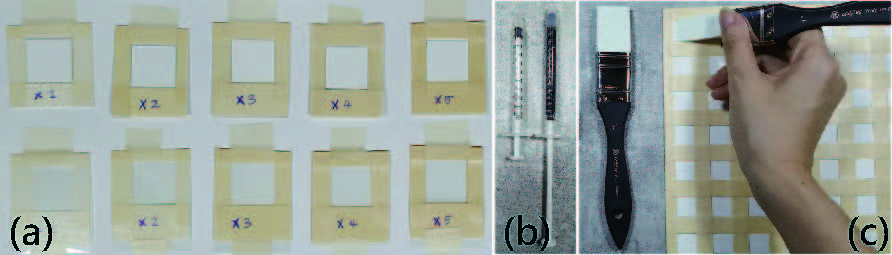}
\caption{The tools for making the watercolor samples. (a) To make samples of different thicknesses, we use masking tape around each grid square to maintain the thickness of the coated pigment and add a layer of masking tape for every extra 0.02 mL; the top row is grids of white paper, and the bottom row is grids of transparent film. (b) The syringes used for measuring pigment quantity. (c) Silicone wide-firm flat brushes. To effectively coat pigments onto the grid, we trim its size to match the grid length.}
\label{fig:tools}       
\end{figure}

In our dataset, we select 13 pigments that are commonly used by painters. These are cadmium red $\rho_1$, alizarin crimson $\rho_2$, burnt sienna $\rho_3$, lemon yellow $\rho_4$, cadmium yellow $\rho_5$, raw sienna $\rho_6$, sap green $\rho_7$, cerulean blue $\rho_8$, cobalt blue $\rho_9$, ultramarine $\rho_{10}$, prussian blue $\rho_{11}$, ivory black $\rho_{12}$, and chinese white $\rho_{13}$; the brand of watercolor pigments is Winsor and Newton \cite{Winsor_2018b}. The 13 pigments by RGB are shown in Fig. \ref{fig:13pigments}. In the initial state of our experiment, we test-measured few pigments in 16 different quantities, starting from 0.01 mL to 0.16 mL with 0.01 ml as an interval. The results of the testing showed that wavelength becomes too similar to establish meaningful sample. Thus, 0.11 mL and three more quantities between 0.12 mL to 0.16 mL are omitted for the actual data collection. In other words, 12 quantities are sampled for each pigment at the end: 0.0l mL, 0.02 mL, ..., 0.10 mL, 0.12 mL and 0.16 mL. We coat pigments onto two substrates: grids of white paper (Canson Ca grain, its transmittance is as low as 0~2.78$\%$ and has an average value 1.79) and transparent film; the size of each grid is 1.5cm $\times$ 1.5cm. Fig. \ref{fig:tools} shows the substrates, measuring tool, and a brush.

%
%
%
%
%

\begin{figure*}[htbp]
  \includegraphics[width=1.0\textwidth]{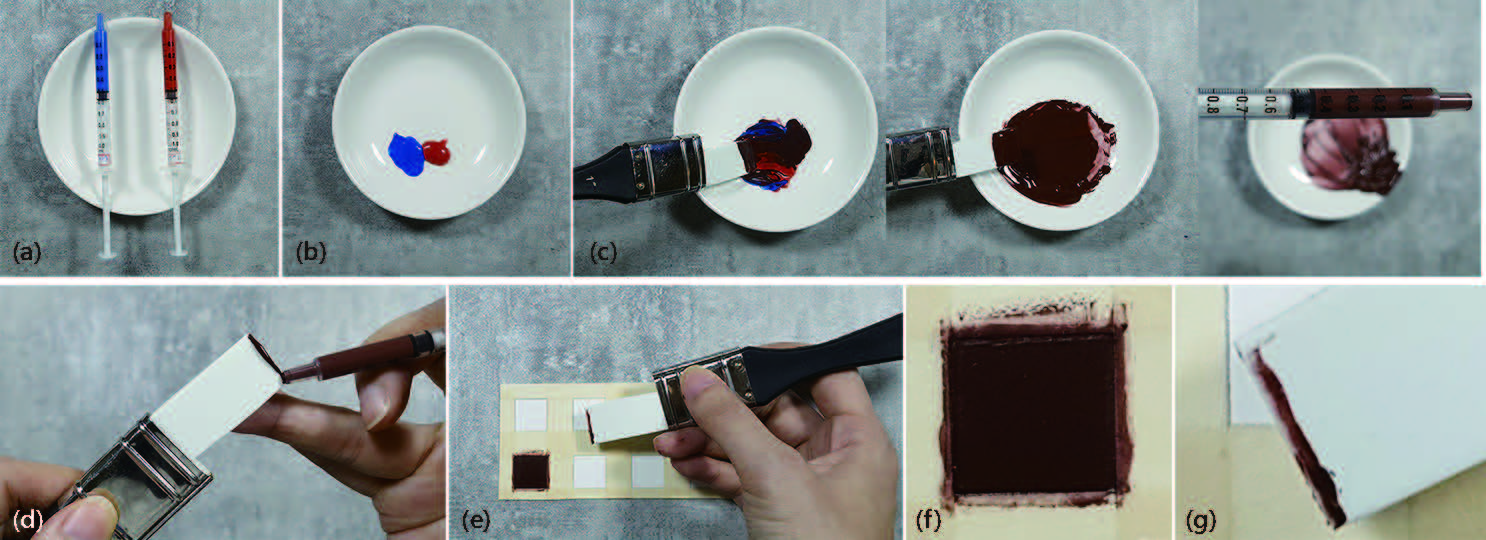}
\caption{A mixing pigment and pigment coating process. (a) Two primary pigments, $\rho_1$ and $\rho_8$, stored in separate syringes of 1 mL capacity. (b) Pumping out 0.5 mL of each pigment for color mixing, resulting in a dark brown, not purple. (c) the mixture result is depositing to a new syringe. (d) Applying 0.02 mL of the mixture on the brush, (e) completing a pigment sample by coating the pigment in a sample grid.  Fig (f) and (g) shows the stray pigments near the sample grid and on the brush in zoom.}
\label{fig:coatedonpaper}       
\end{figure*}

\begin{figure}[htbp]
\centering
  \includegraphics[width = 0.9\textwidth]{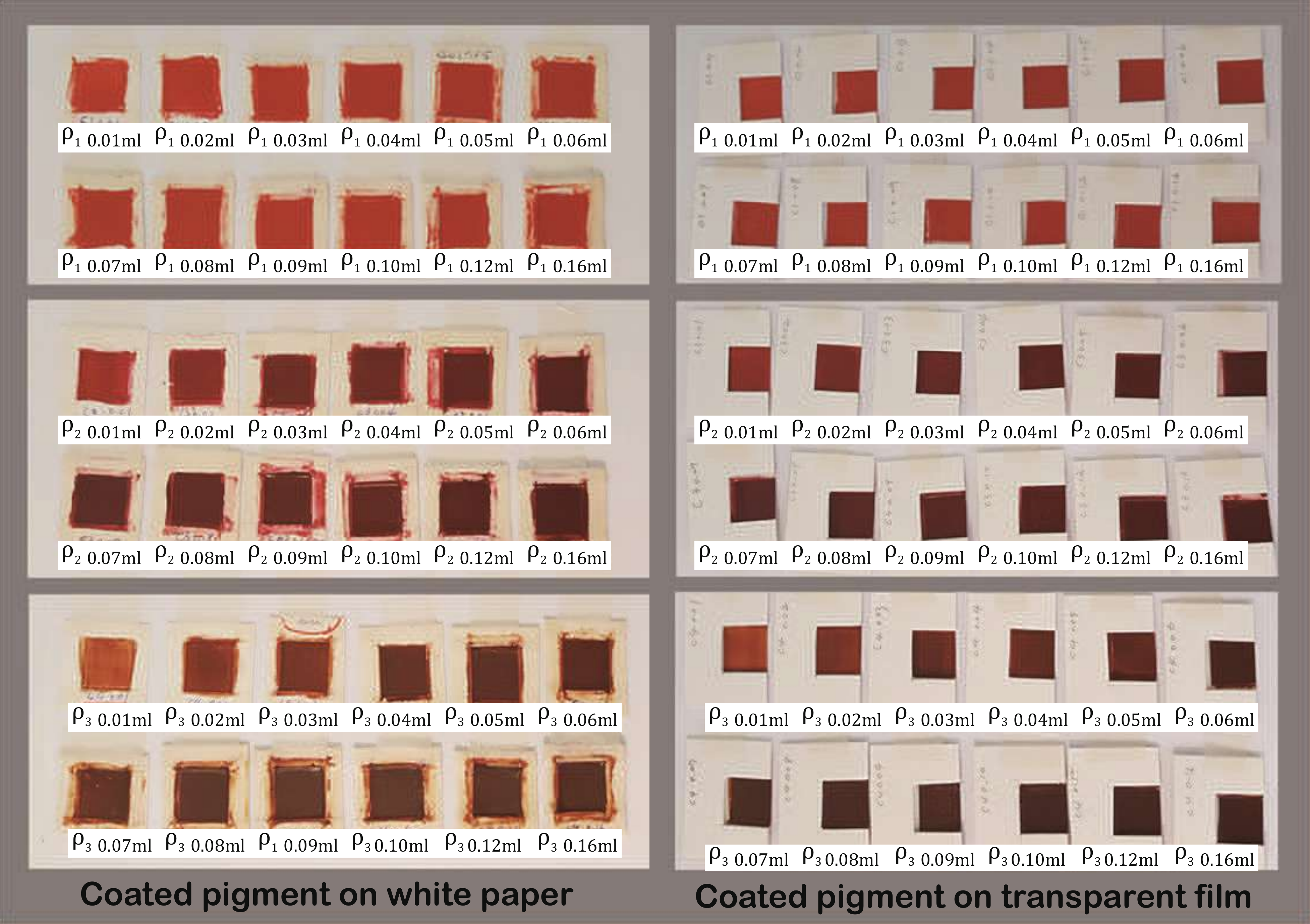}
\caption{Three examples of  primary pigments are coated on white paper and transparent film. The left side is a photo of the coated pigments on white paper; the right side is a photo of the pigments on transparent film. From top to bottom are 12 quantities of cadmium red $\rho_1$, alizarin crimson $\rho_2$, and burnt sienna $\rho_3$. Each grid has the size of 1.5cm $\times$ 1.5cm.}
\label{fig:2coatpigment}       
\end{figure}

\begin{figure}[htbp]
\centering
  \includegraphics[width = 0.5\textwidth]{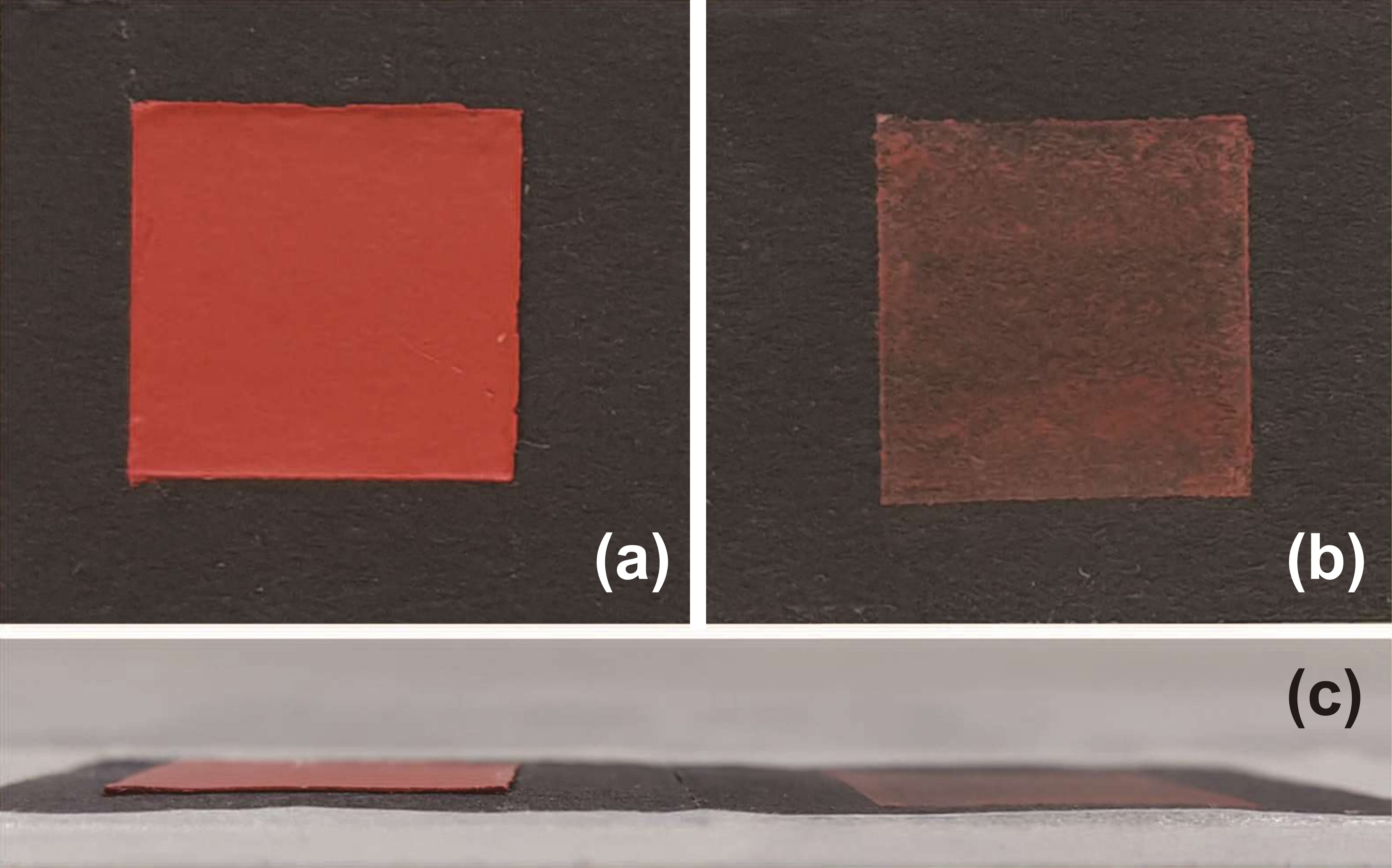}
\caption{Evaluation for the stray pigment occurred during the coating process. We recreate the sample making process for 0.05 mL of cadmium red $\rho_1$, a relatively bright pigment, onto a black-colored substrate. The contrast helps better observing the amount of stray pigment. (a) The initial grid; 0.05 mL of cadmium red $\rho_1$ coated on black sample grid, with the masking tape removed (b) the recollected grid; a new grid coated only with the stray pigments occurred(on brush and masking tape) in the initial coating process. With all stray pigments we recollected, it is still not sufficient to form an even layer on a grid. (c) Viewing the two grids from a near-horizontal angle. The recollected grid shows close to no thickness.}
\label{fig:DataLoss}       
\end{figure}

To obtain more accurate pigment samples, we use a silicone wide-firm flat brush, which is a pliable flat scraper, to replace a traditional brush which tends to trap much pigment residues. Such silicone brush also enables us to coat the layers as evenly as possible, which is a critical factor in data accuracy. In addition, we use a syringe with a 1 mL capacity to measure pigment quantities and masking tape around each grid to maintain the thickness of the coated pigment. The process of coating is shown in Fig. \ref{fig:coatedonpaper}, and the samples of three examples of primary pigments are shown in Fig. \ref{fig:2coatpigment}.  To evaluate the loss pigments occurred during the coating process, we recollect those stray pigments on a black sheet of paper. Fig \ref{fig:DataLoss} illustrate that the loss of pigments with our setup is minimal. 

%
%
%

%

\begin{figure}[hbtp]
\centering
  \includegraphics[width=0.5\textwidth]{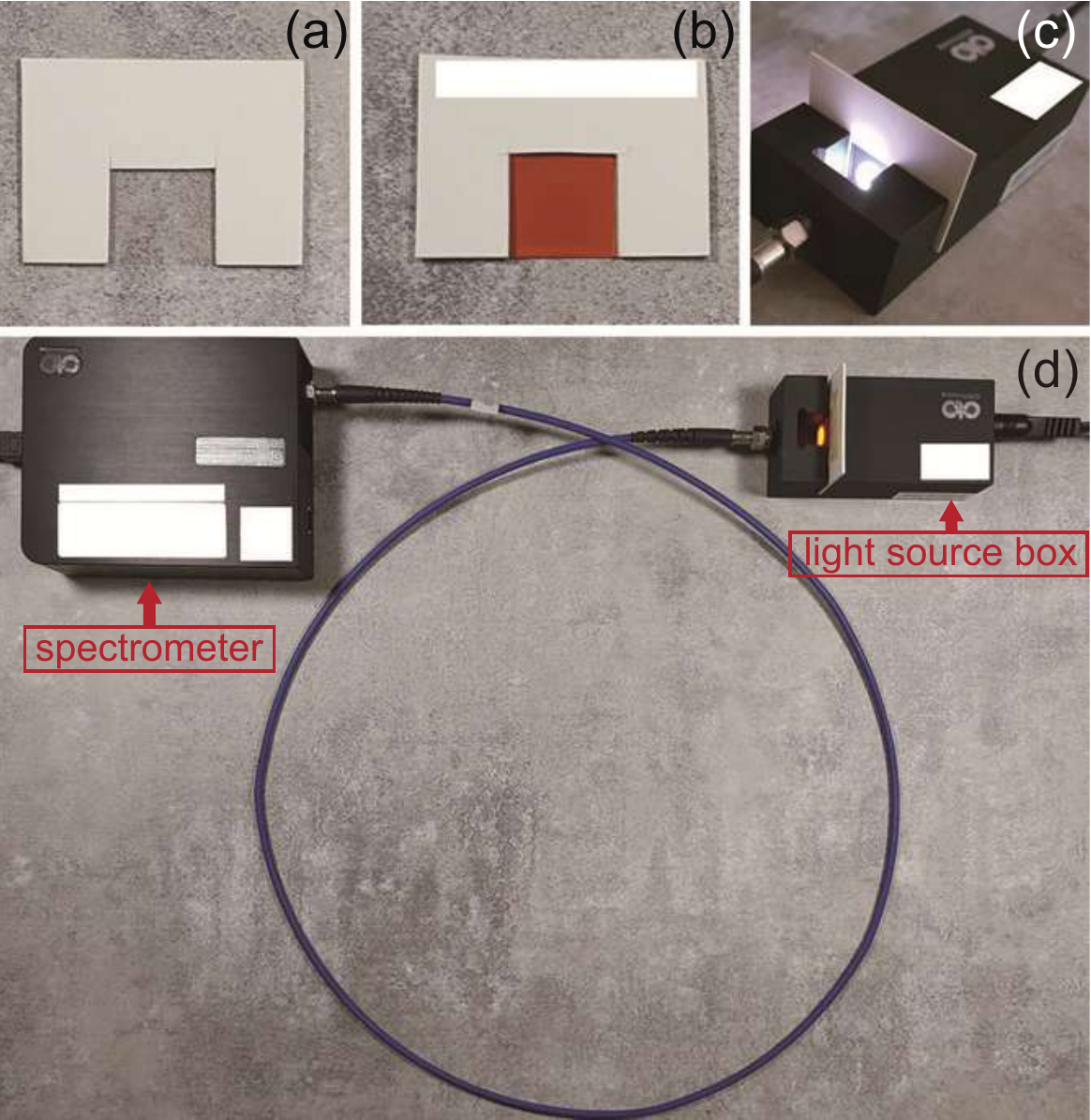}
\caption{The measuring workflow of relative transmittance of pigment sample. (a) using transparent film as reference spectrum for calibrating spectrometer to get relative transmittance. (b) a pigment sample being pigment coated on transparent film (c) the measured sample, namely transparent film (a), is mounted tightly onto the aperture of the light source in light source box module. (d) the setup of the transmittance of measuring pigment sample. The setup of measuring transmittance demonstrate how measurements was taken and not just a showcase of the devices, except that the granite-colored countertop is not the exact location. It was done on a white-colored table.}
\label{fig:T_measured}       
\end{figure}

\begin{figure}[t]
\centering
  \includegraphics[width=0.7\textwidth]{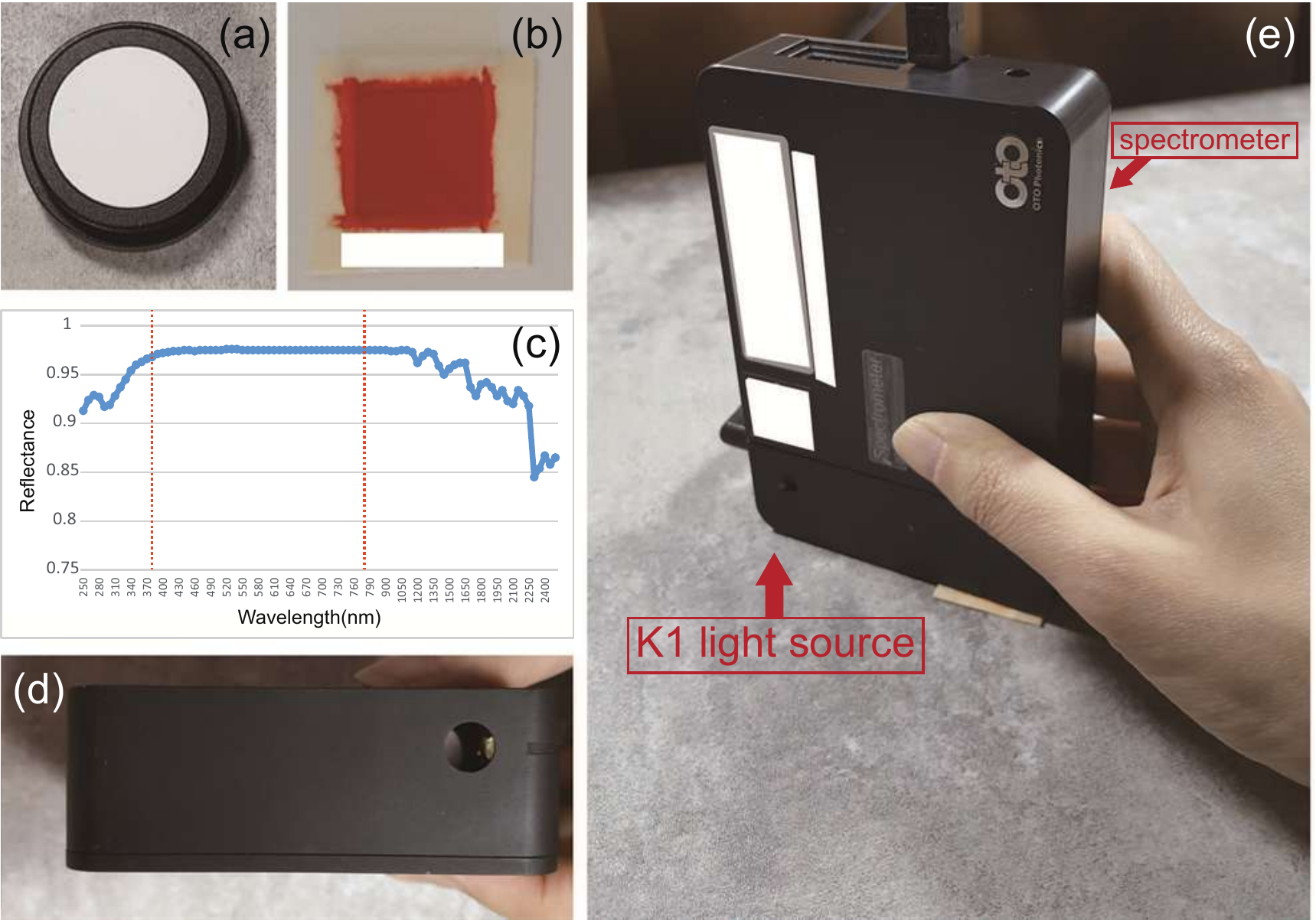}
\caption{The measuring workflow of absolute reflectance of pigment sample. (a) a standard white reflectance for calibrating spectrometer to get absolute reflectance. (b) pigment sample being  pigment coated onto white paper. (c) the reflectance of the standard reflectance which is close to 98$\%$ from 380nm to 780nm. The x-axis is wavelength, and the y-axis is reflectance.(d) a measurement aperture of the reflectance on K1 light source. (e) a setup of the reflectance of measuring pigment sample. The setup of measuring reflectance demonstrate how measurements was taken and not just a showcase of the devices, except that the granite-colored countertop isn’t the exact location. It was done on a white-colored table.}
\label{fig:R_measured}       
\end{figure}

For measuring our samples, we use an SD1220 spectrometer \cite{OTO_2018}, which can detect wavelength ranging from 380 to 780 nm. Two configurable modules containing the same light source, light source box module and K1 light source module, are mounted to the spectrometer, which enable us to measure the transmittance and reflectance, respective, of our watercolor pigment samples. For receiving the spectrum of transmittance and reflectance, the positions of their light sources modules have different arrangements. To get transmittance, the light source box module is attached to the spectrometer through an optical fiber connector, as shown in Fig. \ref{fig:T_measured} (d); on the other hand, the K1 light source module measures reflectance by attaching directly onto the spectrometer, as seen in Fig. \ref{fig:R_measured} (e). Fig.\ref{fig:R_measured} (d) shows the aperture (both light source and measuring) of the K1 light source module. The apertures on both modules have the same size being a circle with a diameter of 0.8 cm. Details of how to operate the devices for measuring our watercolor pigment samples will be introduced in the section below.

In the study, we use a relative transmittance to denote the transmittance of primary pigments. Measuring steps are as follows: 

\begin{enumerate}

\item Setup the devices. We attach the spectrometer and the light source box module together with a fiber optic connector as seen Fig. \ref{fig:T_measured} (d); once complete, connecting the setup to computer. Next, we startup an OTO software SpectraSmart. Then, we choose transmittance measurement and set a required parameter of a peak value around 40,000.

\item Establishing the reference spectrum. We place a transparent film tightly against the light source aperture
in light source box module for calibration. The calibration transparent film is shown in Fig. \ref{fig:T_measured} (a), and the complete setup for calibration is shown in Fig. \ref{fig:T_measured} (c); This step establishes the reference spectrum.

\item Establishing the dark spectrum.. We take away the transparent film and turn off the light source. This setting, with no film and no light source, establishes as the dark spectrum.

\item Measuring the sample. Turning on the light source again and start placing the pigment samples (Fig. \ref{fig:T_measured} (b)) to light source box module, the transmittance of pigment samples are then measured and recorded. 

\item[$\ast$]  The transmittance obtained through this workflow only contain that of the pigment samples. In other words, the transmittance of the transparent film is removed. Hence,  the result of this measuring method called the relative transmittance of pigment samples.

\end{enumerate}

%

\FloatBarrier
We measure the absolute reflectance of primary and mixture pigment samples (pigments coated on white paper). The measuring workflow is as follows:

\begin{enumerate}

\item Setup the devices. Attaching the K1 light source module directly onto spectrometer, as seen Fig. \ref{fig:R_measured} (e), then connect the setup to computer. Next, we startup an OTO software SpectraSmart and select to measure reflectance; then, we input the standard white reflectance curve and set a required parameter of a peak value around 40,000.

\item Establishing the reference spectrum. We use the standard white reflectance (as seen Fig. \ref{fig:R_measured}(a) ) as the reference spectrum.

\item Establishing the dark spectrum. We take away the standard white reflectance and also turn off the K1 light source module. We establishes this setting no light source as the dark spectrum.

\item Measuring the sample. Turning the K1 light source module back on again and start placing the pigment samples (Fig. \ref{fig:R_measured}(b)) on a flat surface, with the measuring device pressed vertically and directly onto the pigment samples, as Fig. \ref{fig:R_measured} (e); demonstrates; then, we measure the pigment sample and get its reflectance.

\item[$\ast$] The reflectance obtained through with method is called the absolute reflectance of the measured pigment. 
\end{enumerate}

One notable thing is that, every time when the devices reconnect to the computer or when the software SpectrumSmart is restarted, we must start the workflow from step one again, including all the software setting and calibration. 

\FloatBarrier

\begin{figure}[htbp]
  \centering
  \includegraphics[width=0.9\textwidth]{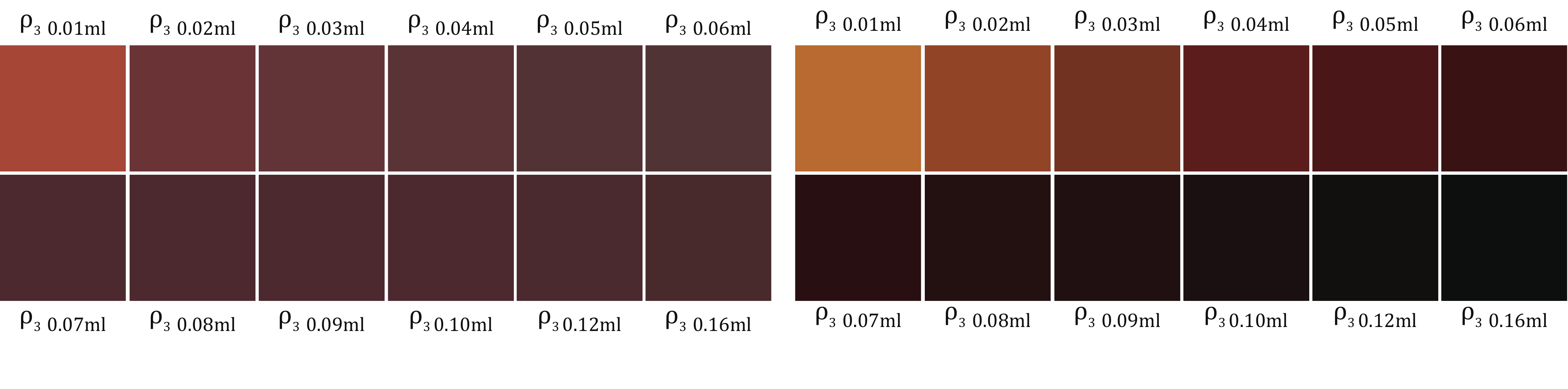}
\caption{RGB colors of the absolute reflectance and relative transmittance of burnt sienna $\rho_3$. The left half is the absolute reflectance spectrum converted to RGB color; the right half is the relative transmittance spectrum converted to RGB color.}
\label{fig:P3_RT}       
\end{figure}

\begin{figure}[htbp]
\centering
  \includegraphics[width=0.7\textwidth]{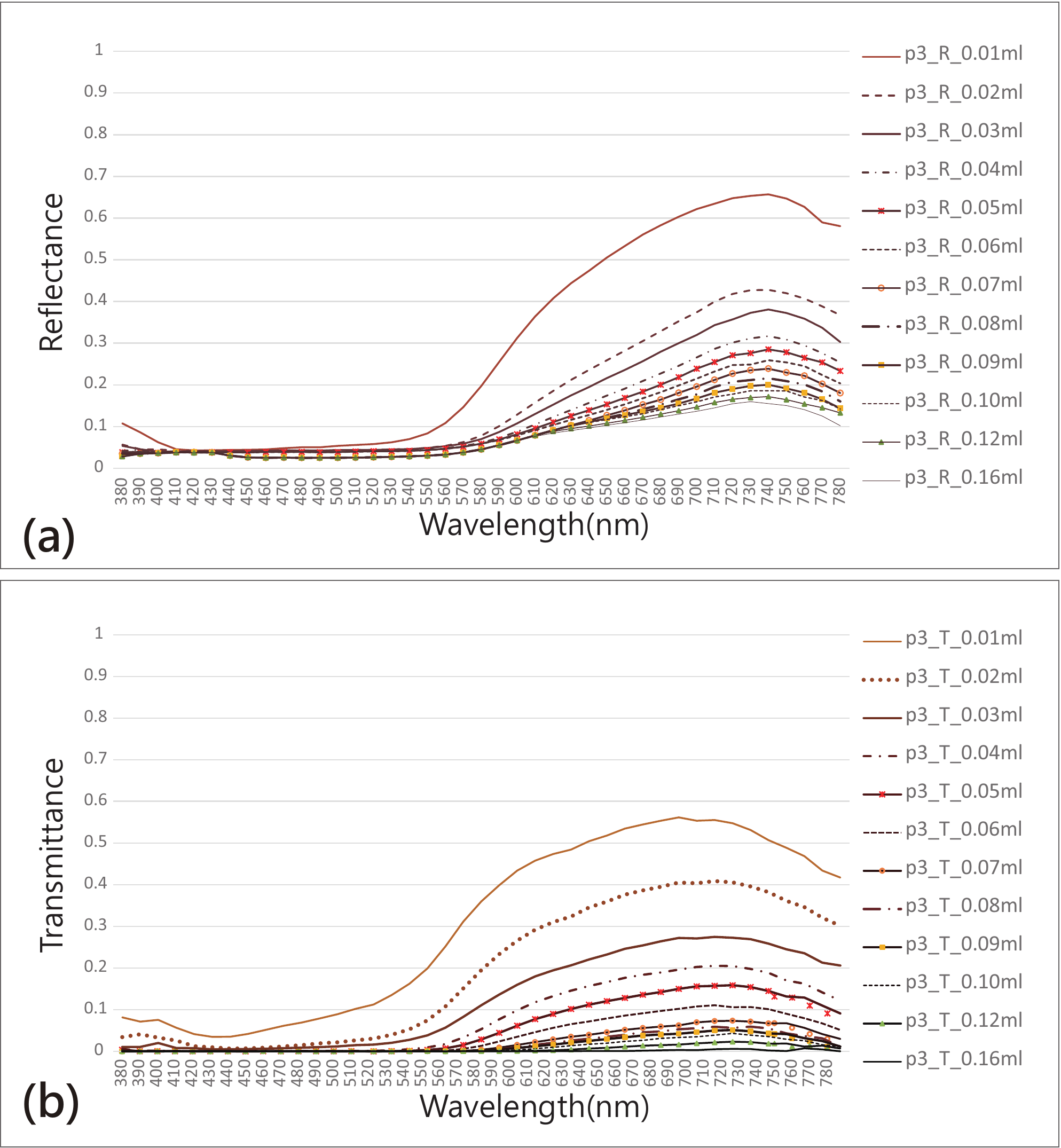}
\caption{The absolute reflectance and relative transmittance of burnt sienna $\rho_3$. The x-axis is wavelength, and the y-axis is reflectance/transmittance. }
\label{fig:P3_RT_curve}       
\end{figure}

In our study, each relative transmittance and absolute reflectance is denoted 41 nm samples which is taken per 10nm pitches as a sample from 380 to 780 nm. Finally, we measure 156 relative transmittance and 156 absolute reflectance from our making samples which are 13 primary pigments with 12 quantities coated on transparent film and white paper respectively and 780 absolute reflectance of two primary pigment mixture in total. Fig. \ref{fig:P3_RT} shows absolute reflectance and relative transmittance data collected from 12 quantity samples of burnt sienna $\rho_3$ is represented in RGB, whereas Fig. \ref{fig:P3_RT_curve} depicts the sample nm using line charts.

%
%
%

\FloatBarrier
\subsubsection{Labeling data to construct NTU WPSM dataset}
\label{labelData}

After making and measuring pigment samples we obtain those data: the relative transmittance and absolute reflectance of the primary pigments, the absolute reflectance of two primary pigment mixtures and the white paper, and the quantity of the primary pigments. Those data are labeled to train SWPM prediction model.

\begin{figure*}[htbp]
  \includegraphics[width=1.0\textwidth]{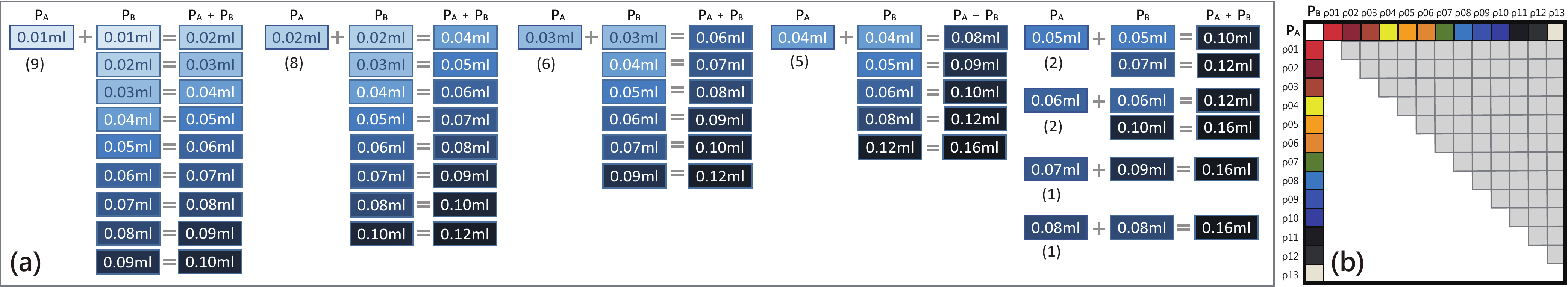}
\caption{Two types of labeled data in our dataset. (a) Data Type I.. Each pigment has 34 combinations to label; thus, the total number of labeled data is 13 $\times$ 34. (b) Data Type M.. Two quantities are selected to mix, which can be labeled to 78 mixed combinations; we select 10 pairs of various pigment quantities; therefore, the total number of labeled data is 10 $\times$ 78. } 
\label{fig:dataset_label}       
\end{figure*}

There are two labeled types of color mixing in NTU WPSM dataset: Data Type I. is collected from adding the same primary pigments to itself, namely increasing the thickness/quantities of a primary pigment. ( see Fig. \ref{fig:dataset_label} (a)) and Data Type M. is collected by mixing two different primary pigments with three mixing ratio (1;1, 1:2, and 2:1) in different quantities.(see Fig. \ref{fig:dataset_label} (b)). The combination ratios employed in Data Type M. are as follows:

\begin{enumerate}
\item $P_A$ : $P_B$ = 50$\%$ : 50$\%$, which is equivalent to 0.01 mL and 0.01 mL,  0.02 mL and 0.02 mL, 0.04 mL and 0.04 mL, and 0.08 mL and 0.08 mL.
\item  $P_A$ : $P_B$ = 33.3$\%$ : 66.6$\%$, which is equivalent to 0.01 mL and 0.02 mL, 0.02 mL and 0.04 mL, and 0.04 mL and 0.08 mL.
\item  $P_A$ : $P_B$ = 66.6$\%$ : 33.3$\%$, which is equivalent to 0.02 mL and 0.01 mL, 0.04 mL and 0.02 mL, and 0.08 mL and 0.04 mL.
\end{enumerate}
 
\vspace{0.1cm}   
%
 
\vspace{2.5cm} 
In short, NTU WPSM dataset contains 442 data labeled as Data Type I. and 780 data labeled as Data Type M. (Fig. \ref{fig:dataset_label}): each data has 7 components, the relative transmittance and absolute reflectance of two primary pigments ---$T_{\rho_A}$, $T_{\rho_B}$, $Rw_{\rho_A}$ and $Rw_{\rho_B}$--- the absolute reflectance of the white paper, $Rw$, two quantities of the two primary pigments, $Q_{\rho_A}$ and $Q_{\rho_B}$, and the absolute reflectance of the two primary pigment mixtures, $Rw_{mix}$, and the last one is labeled as  an answer of the first six.   

\FloatBarrier

\subsection{SWPM prediction model for color mixing using DNN}

In Table \ref{Table_data}, color mixing is an unordered problem, we must change the order of $P_A$ and $P_B$; therefore, all data are doubled. Then, we take 80$\%$ of the data as training data and 20$\%$ of the data as test data from each type of NTU WPSM dataset, Data type I. and M.. To decide a prediction model of color mixing, at first, we use 20$\%$ data from training data as validation data and the rest data of training set for training. After the model is determined, we use all data of training set to train SWPM prediction model and apply SWPM prediction model to predict mixtures.  

\begin{table}[htbp]
\centering
\caption{Training and test sets}
\label{Table_data}
\begin{tabular}{|c|c|c|c|c|}
\hline
Data        & \multicolumn{2}{c|}{Training set} & \multicolumn{2}{c|}{Test set} \\ \hline
Type        & I.              & M.              & I.            & M.            \\ \hline
$P_A$+$P_B$ & 354             & 624             & 88            & 156           \\ \hline
$P_B$+$P_A$ & 354             & 624             & 88            & 156           \\ \hline
Total       & \multicolumn{2}{c|}{1956}         & \multicolumn{2}{c|}{488}      \\ \hline
\end{tabular}
\end{table}

\begin{figure}[htbp]
  \includegraphics[width=1.0\textwidth]{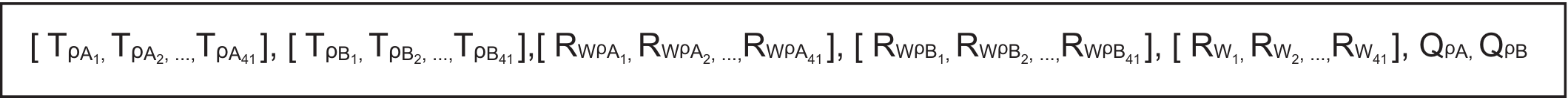}
\caption{Input layer of SWPM prediction model. $T_{\rho_A}$, $T_{\rho_B}$, $Rw_{\rho_A}$, and $Rw_{\rho_B}$ denote the relative transmittance of $P_A$ and $P_B$ and absolute reflectance of $P_A$ and $P_B$ coated onto white paper; $Rw$ is the absolute reflectance of white paper; thus, there are 5 $\times$ 41 features; $Q_{\rho_A}$ and $Q_{\rho_B}$ are the quantities of $P_A$ and $P_B$; thus, the total is 207 features as input data.}
\label{fig:207}       
\end{figure}

In SWPM prediction model, an input layer has 207 features that are the transmittance of two primary pigments ---$T_{\rho_A}$, $T_{\rho_B}$--- the reflectance of two primary pigments coated onto white paper ---$Rw_{\rho_A}$, $Rw_{\rho_B}$--- the reflectance of white paper, $Rw$ (where these data are normalized), and the quantities of the two primary pigments, $Q_{\rho_A}$ and $Q_{\rho_B}$ (as seen Fig. \ref{fig:207}). The output layer is the reflectance of a mixture coated onto white paper, $Rw_{mix}$. $T_{\rho_A}$, $T_{\rho_B}$, $Rw_{\rho_A}$, $Rw_{\rho_B}$, $Rw$, and $Rw_{mix}$ are are a spectrum denoted 41 sampled nm except $Q_{\rho_A}$ and $Q_{\rho_B}$.

%

%
\begin{figure}[htbp]
  \centering
  \includegraphics[width=0.7\textwidth]{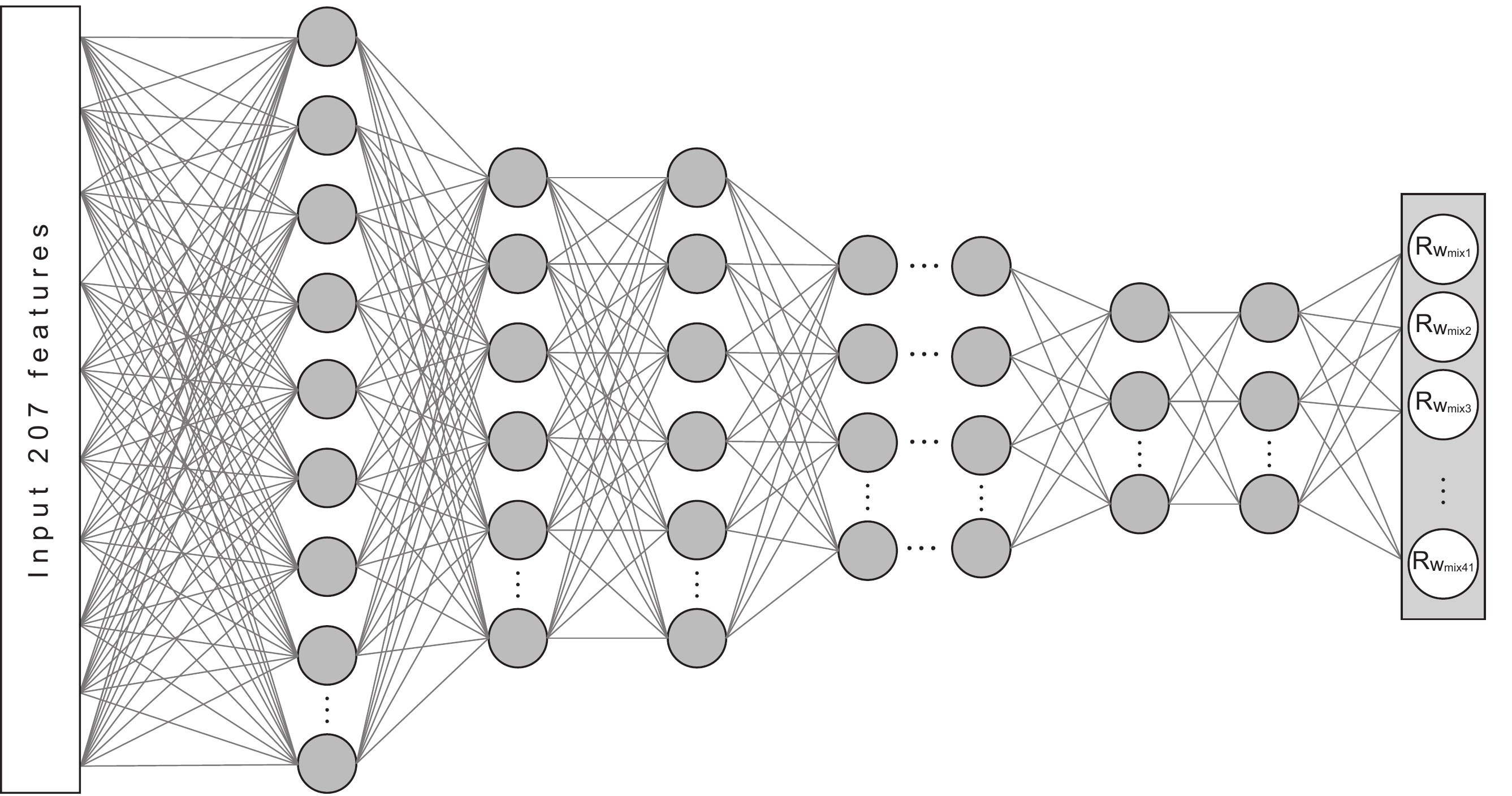}
\caption{Architecture of SWPM prediction model. The model has 15 hidden layers containing 100, 90, 90, 80, 80, 70, 70, 60, 60, 60, 60, 50, 50, 50, and 50 neurons, respectively.}
\label{fig:dnn_207}       
\end{figure}

\vspace{0.5cm}
We use a DNN to train SWPM prediction model, Fig. \ref{fig:dnn_207} presents the SWPM prediction model architecture which has 15 hidden layers containing 100, 90, 90, 80, 80, 70, 70, 60, 60, 60, 60, 50, 50, 50, and 50 neurons, respectively. The output of the model is the reflectance at 41 sampled nm of a mixed pigment coated onto paper, $Rw_{mix}$. We use a sigmoid as the activation function and Adam as our optimizer; its learning rate is 0.001. The loss function uses the mean absolute deviation percentage, mean absolute error (MAE), and standard deviation (SD) without square root (see Eq. \ref{loss_func}); L1-norm and L2-norm are the regularization terms, and they are multiplied by the weights 0.000015 and 0.000003, respectively.   
Finally, 3.7 million epochs are used and we use the Google TenserFlow for implementation \cite{abadi2016tensorflow}.

\begin{equation}\label{loss_func}
 \frac{\sum_{i=1}^{m}\sum_{k=1}^{41}\left | \mathbf{y}_{k}^{i}-\mathbf{\hat{y}}_{k}^{i} \right |}{ \sum_{i=1}^{m}\sum_{k=1}^{41}\left |\mathbf{y}_{k}^{i} \right |}
+\frac{\alpha}{m} \sum_{i=1}^{m}\sum_{k=1}^{41}\left | \mathbf{y}_{k}^{i}-\mathbf{\hat{y}}_{k}^{i}\right | \\
+ \frac{\beta }{m}\sum_{i=1}^{m}\sum_{k=1}^{41} \left (\mathbf{\hat{y}}_{k}^{i}-\mu   \right )^{2}
\end{equation}
where $m$ is the size of the dataset, \textbf{y} is the vector with the true value, and $\hat{\textbf{y}}$ is the vector with the predicted value; each vector has 41 elements, and $\mu$ is the average of \textbf{y}. In addition, $\alpha$ and $\beta$ are the weights of MAE and SD without square root, which are set to 3 and 2, respectively. 
\\
%

\section{Results}
To make a verdict on whether SWPM prediction model is valid for what is created for, which is helping watercolor novices to learn watercolor pigment mixture with precision, we conduct evaluation and comparison through experiments. In our experiments, we display the pigment colors using standard RGB (sRGB) with CIE 1931 2$^\circ$ observer and standard illuminant D65. In Sect. \ref{CompDelta}, we use a color difference formula, $\Delta E^{*}_{ab}$, to evaluate the accuracy of our prediction results. In Sect. \ref{CompKM}, we compare our prediction results with the results obtained using two-constant KM theory. 

\subsection{Using the standard threshold to evaluate color difference}
\label{CompDelta}
\begin{figure}[htbp]
\centering
  \includegraphics[width=0.55\textwidth]{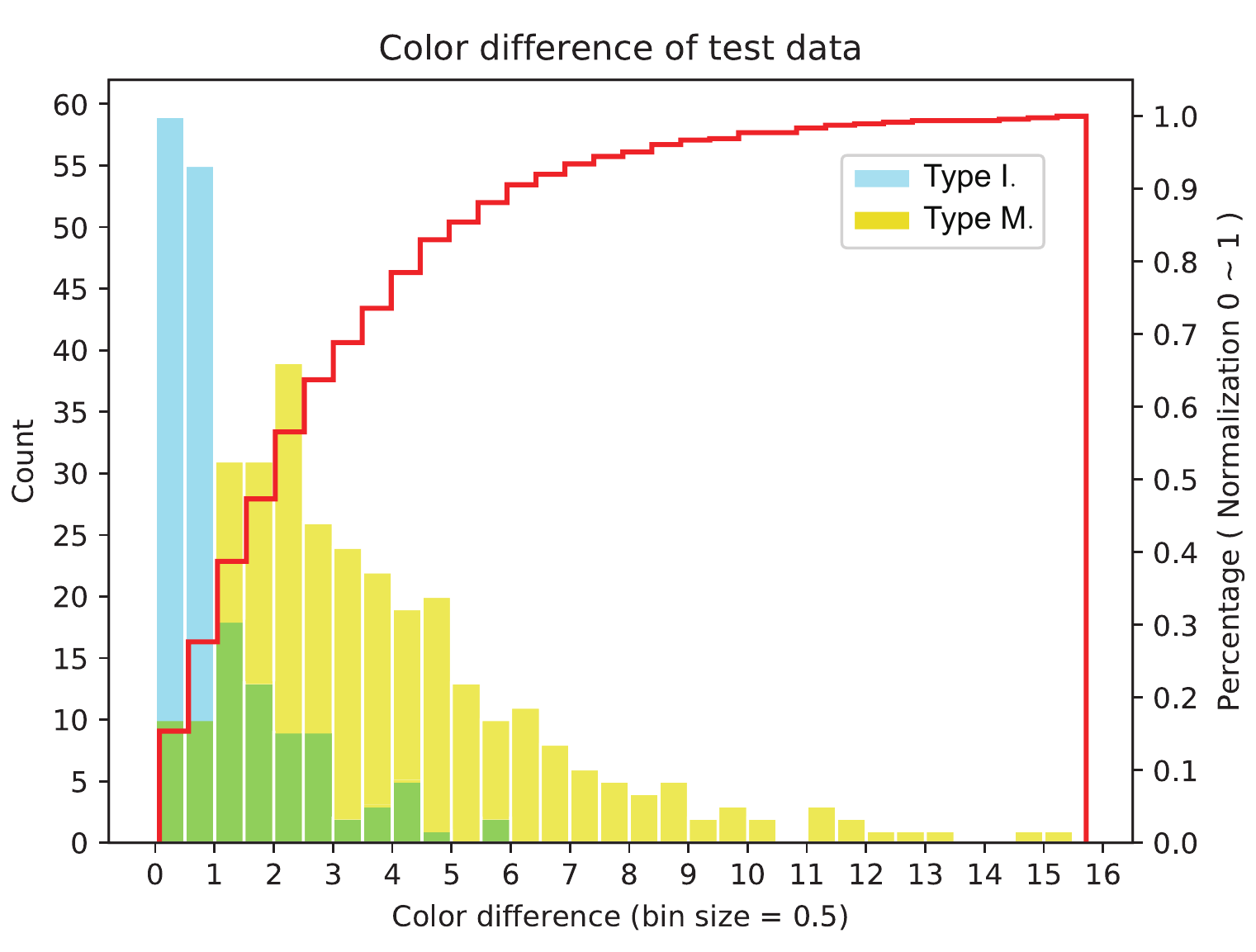}
\caption{Color difference distribution and cumulative distribution function (CDF) of the test data. The left y-axis indicates the cumulative count of color differences, the right y-axis represents the percentage of CDF, and the x-axis denotes color difference values. The cyan bins denote the color difference distribution of Data Type I., and the yellow bins display the color difference distribution of Data Type M.. In addition, the red stepped line is the CDF of Data Type I. and M.. Green area means the color bins of Data Type I. and M. that are overlapped.}
\label{fig:DeltaEcount}       
\end{figure}

\begin{figure*}[tbp]
\centering
  \includegraphics[width=0.7\textwidth]{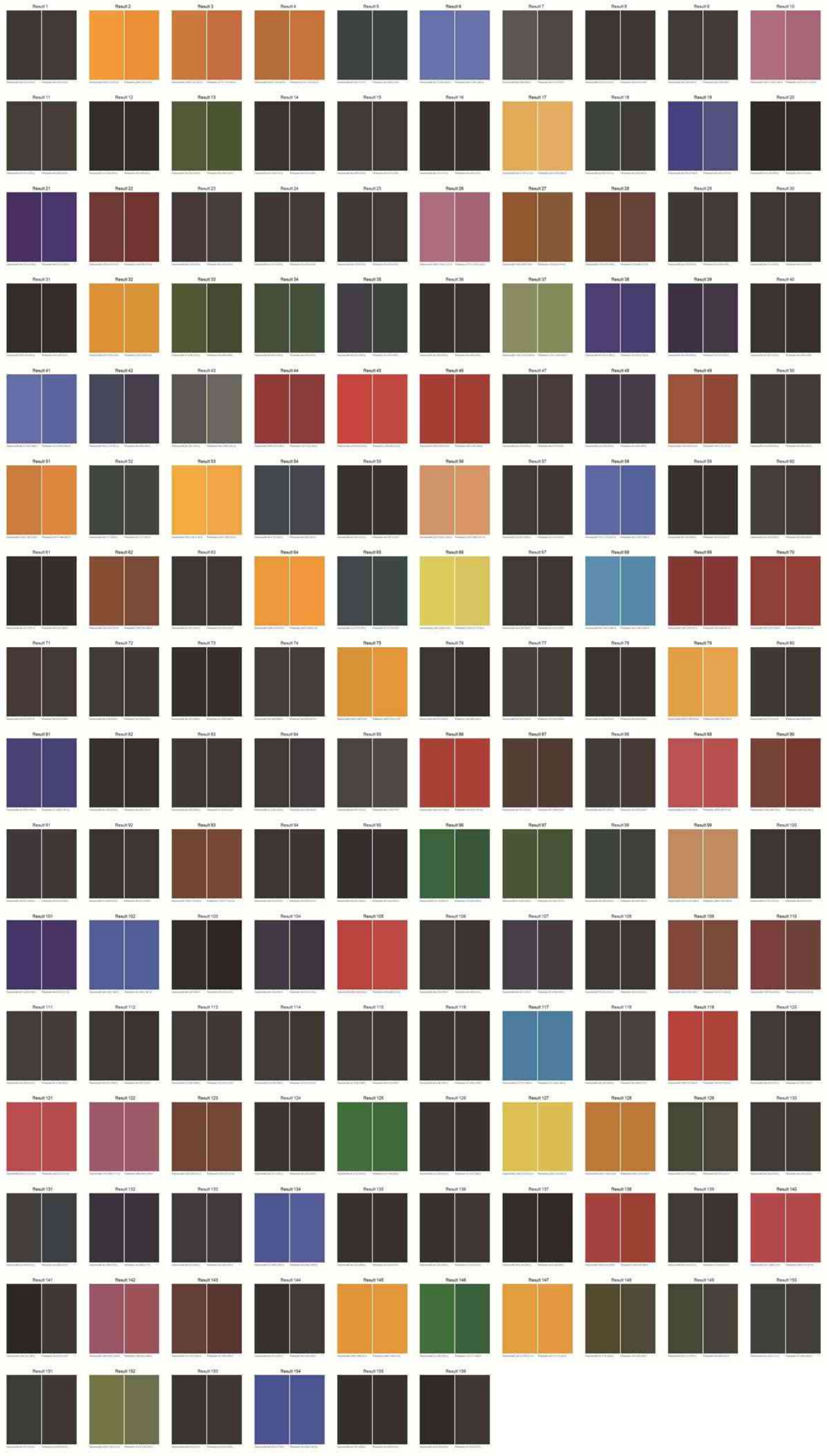}
\caption{RGB colors of the prediction results of two primary pigments for mixing, namely the test data of Data Type M. in our dataset. Each square has two color blocks to represent a result; the left block is the ground truth, the color of the real mixed pigment, and the right block is the result of our prediction. The mixture colors of $P_A$ plus $P_B$ are shown.}
\label{fig:156RGB}       
\end{figure*}

\begin{figure*}[tbp]
  \includegraphics[width=1.0\textwidth]{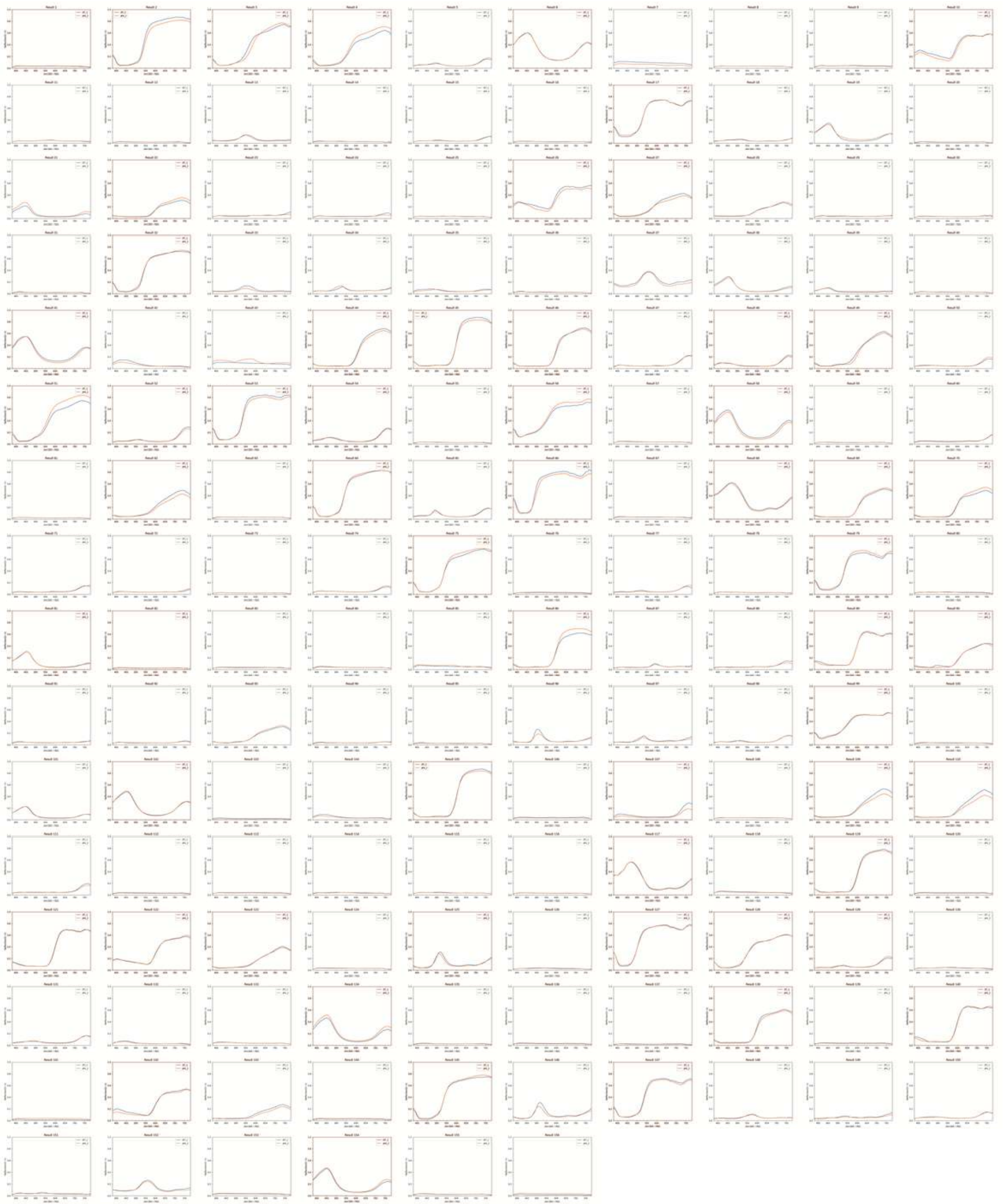}
\caption{The curves of the prediction results of two different primary pigments for mixing, namely the test data of Data Type M. in our dataset. Each image has two curves; the cyan curve is the reflectance of the ground truth, the orange curve is the reflectance of a mixture of $P_A$ plus $P_B$.}
\label{fig:156Curve}       
\end{figure*}

We use a color difference formula $\Delta E^{*}_{ab}$ as shown in Eq. \ref{DeltaEq}, to evaluate our prediction results of the test set. The calculation shows that in comparing our prediction outcomes against the ground truth, close to 83$\%$ of the test set registered a color distance of less than 5, which is below the threshold for an observer to determine that the colors in comparison as two different colors \cite{mokrzycki2011colour}.  The evaluation is visualized in Fig. \ref{fig:DeltaEcount}. In addition, Fig. \ref{fig:DeltaEcount} displays the distribution of two types of test set. Data Type I. has less color difference overall, with the color distance locate mostly between 0 to 3; the color difference distribution for Data Type M. of the test data is mostly distributed from 0 to 5. For a clear illustration,  the prediction results for Data Type M. test set is displayed alongside to the ground truth in both the RGB colors and reflectance curves in Fig. \ref{fig:156RGB} and \ref{fig:156Curve}, respectively.

\begin{equation}\label{DeltaEq}
\Delta E^{*}_{ab} = \sqrt{(\Delta L^{*})^{2}+(\Delta a^{*})^{2}+(\Delta b^{*})^{2}}
\end{equation}
where $\Delta L^{*}$ is a lightness difference, $\Delta a^{*}$ is the red/green difference, and $\Delta b^{*}$ is the yellow/blue difference.

\begin{figure}[tbp]
\centering
  \includegraphics[width=0.7\textwidth]{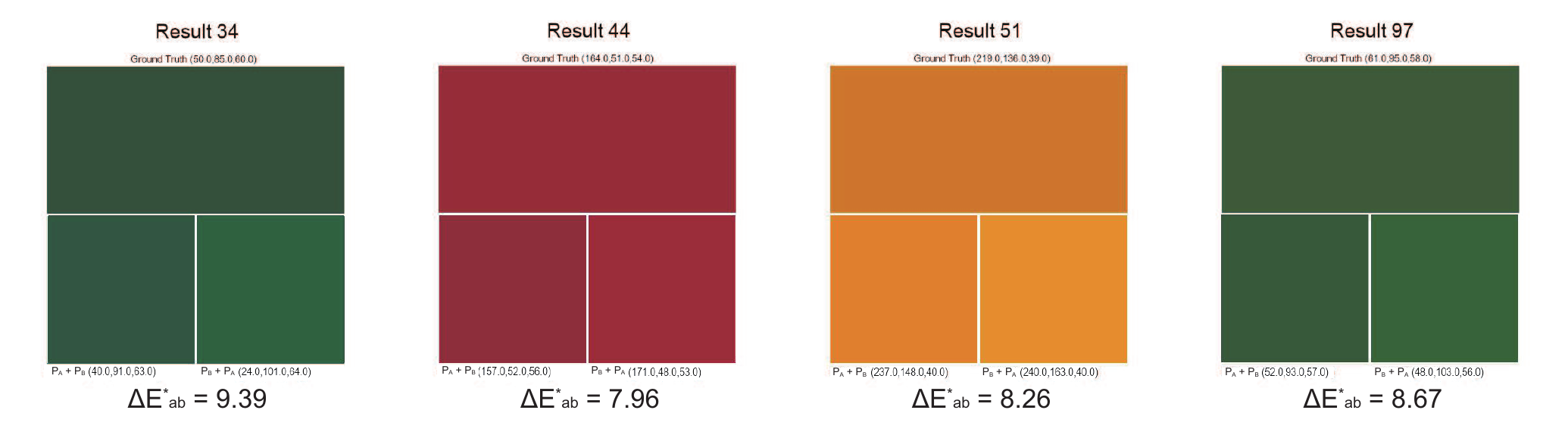}
  \caption{Asymmetry of the prediction model of color mixing. Each square has three color blocks; the top rectangle is the ground truth, the bottom square on the left side is $P_A$ plus $P_B$ of the prediction result, and the bottom-right is $P_B$ plus $P_A$ of the prediction result.}
  \label{fig:symmetry}
\end{figure}

The color difference formula is also used on evaluating the severity level for our symmetry issue, which had occurred as a byproduct for using DNN for training our model. The symmetry issue is where the prediction for $P_A$ plus $P_B$ is not equivalent to that of $P_B$ plus $P_A$ even if the quantity used in mixture are identical. We demonstrate few most severe cases where asymmetry causes the difference in the predicting color in Fig. \ref{fig:symmetry}. Among those cases, the asymmetric prediction in Result 39 has the highest color distance at $\Delta E^{*}_{ab}$ = 9.39, yet one of the two prediction is still fairly close to the ground truth.  
%

\FloatBarrier
\subsection{Comparison of the results of two-constant KM theory}
\label{CompKM}

\begin{figure}[htbp]
\centering
  \includegraphics[width = 0.7\textwidth]{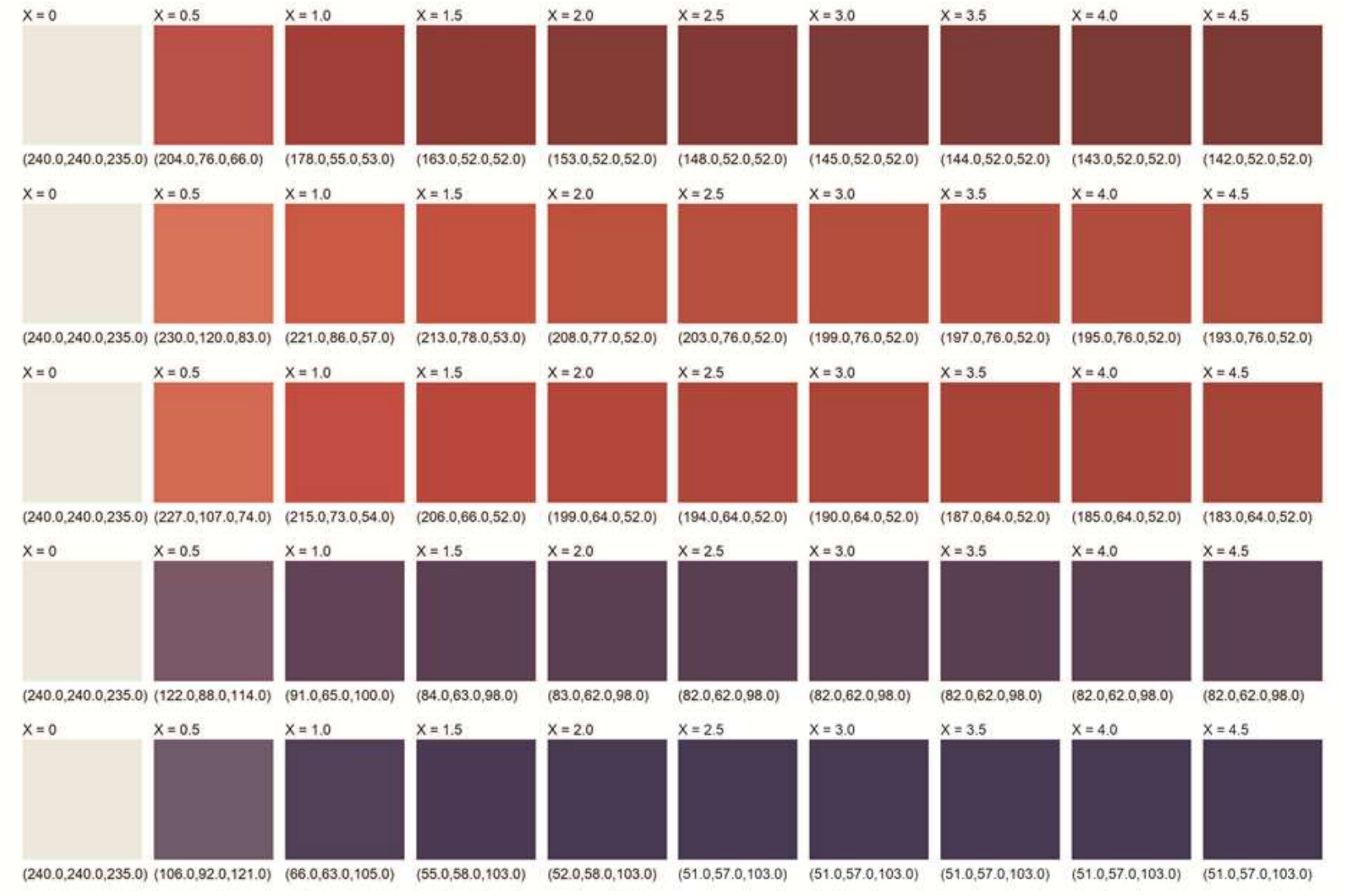}
\caption{5 examples of pigment mixture results in different thickness level X, produced by the two constant KM theory. X = 0 represent that no pigments is being coated onto paper, thus shown as white. These are the same first 5 mixture examples that can be found in Fig. 4. 6, counting from top left to bottom right.}
\label{fig:X_KM}       
\end{figure}

As another way to assess our result, we compare our method to that of the two-constant KM theory. To obtain the absorption, $K$, and scattering, $S$, of watercolor pigments, we use Curtis et al. method \cite{curtis1997computer}, which is a simple inversion of the KM equations. First, we make samples from 0.01 mL of primary pigments being each coated on the white paper grid and the black paper grid. Next, we measure the absolute reflectance of the samples, $R_{W\rho}$ and $R_{B\rho}$, and the absolute reflectance of the white paper grid itself, $R_W$ ($R_{W\rho}$ and $R_W$ are already in NTU WPSM dataset so we only do the samples of pigments coated on black paper grid and measuring the sample, $R_{B\rho}$). Then, we convert $R_{W\rho}$, $R_{B\rho}$, and $R_W$ to RGB values and normalize the values between 0 and 1 (Curtis et al. use an RGB value to replace with the reflectance). After that, using Eq. \ref{97_S} to Eq. \ref{97_a_b}, $K$ and $S$ are then computed into absorption and scattering \cite{duncan1940colour} of the supposed pigment mixture, $K_{mix}$ and $S_{mix}$, using Eq. \ref{40_mixed_KM}. Finally, the mixed reflectance is calculated \cite{kubelka1948new} using Eq. \ref{48_comp_R} and Eq. \ref{48_comp_a_b}.

\begin{equation}\label{97_S}
S = \frac{1}{b}\cdot \coth^{-1}\left ( \frac{b^2-(a-R_{W\rho})(a-1)}{b(1-R_{W\rho})} \right )
\end{equation}

\begin{equation}\label{97_K}
K = S(a-1)
\end{equation}

\begin{equation}\label{97_a_b}
a = \frac{1}{2}(R_{W\rho}+\frac{R_{B\rho}-R_{W\rho}+1}{R_{B\rho}}),\quad
b = \sqrt{a^{2}-1}
\end{equation}
\\
where $R_{W\rho}$ denotes the reflectance of pigment coated on white paper; where $R_{B\rho}$ denotes the reflectance of pigment coated on black paper.  
\\
\begin{equation}\label{40_mixed_KM}
K_{mix}=\sum_{n}^{i = 1}K_{i}c_{i},\quad
S_{mix}=\sum_{n}^{i = 1}S_{i}c_{i}
\end{equation}
\\
where $K_i$ and $S_i$ denote the absorption coefficients and scattering coefficients of each pigment, $c_i$ denotes a proportion of each pigment; $K_{mix}$ and $S_{mix}$ is the calculated result, which is a mixed pigment; in this case, we use two pigments for mixing, $i = 2$, and $c_1$ and $c_2$ are both 0.5.
\\
\begin{equation}\label{48_comp_R}
R = \frac{1-R_{W}(a-b\coth (bS_{mix}X))}{a + b \coth(bS_{mix}X)-R_{W}}
\end{equation}

\begin{equation}\label{48_comp_a_b}
a = 1+ \frac{K_{mix}}{S_{mix}},\quad
b = \sqrt{a^{2}-1}
\end{equation}
\\
where $X$ denotes the thickness of the watercolor pigments. At the end of the equations above, the thickness parameter, $X$, can be adjusted to obtain different results. The results darken as $X$ increases, which is illustrated in Fig. \ref{fig:X_KM} with the $X$ at 10 different levels. 

\begin{figure}[htbp]
\centering
  \includegraphics[width = 0.7\textwidth]{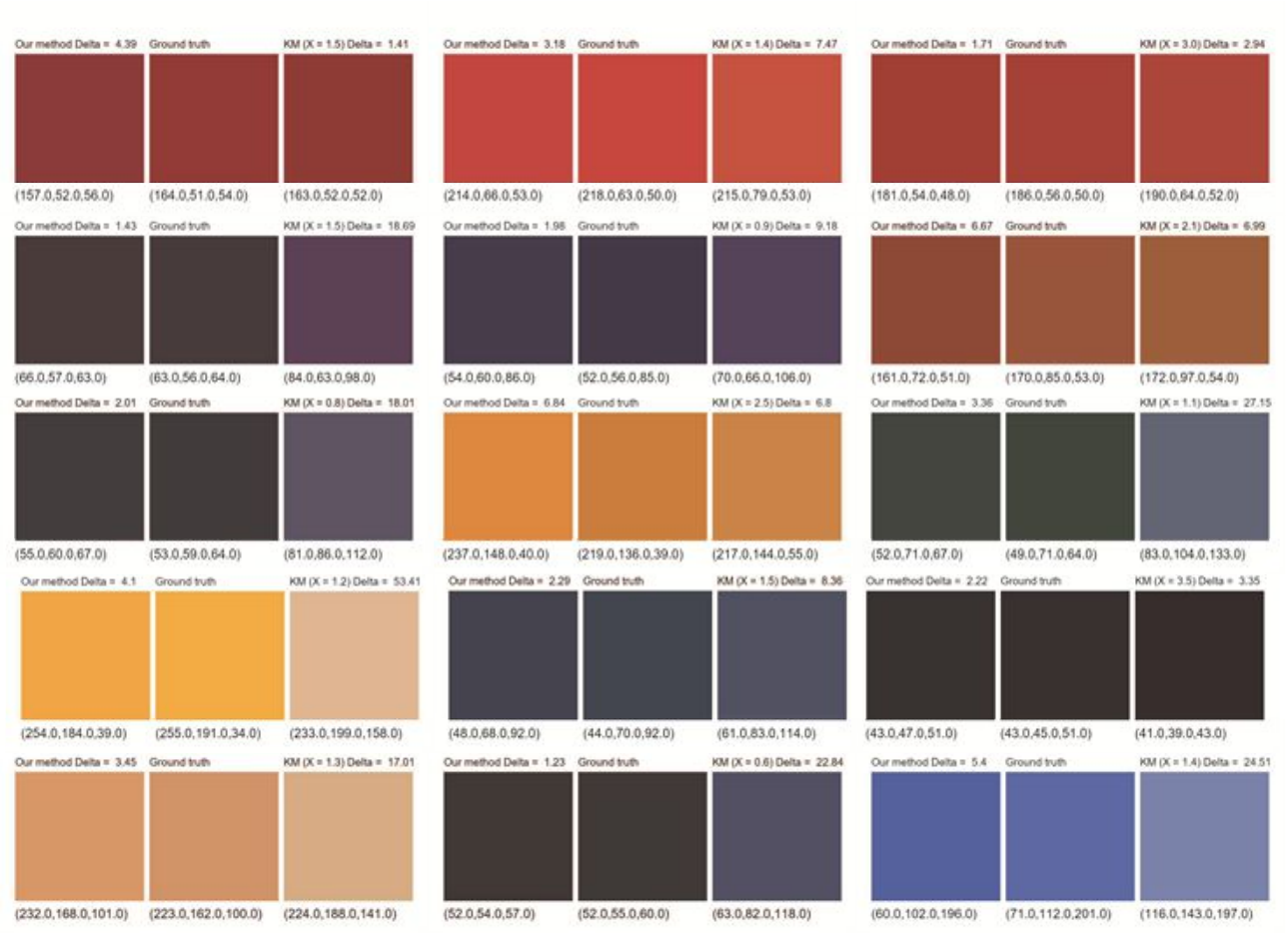}
\caption{ 15 cases of tri-directional comparison between our SWPM prediction results, ground truth, and two constant KM theory. 
 The placement order for the three blocks from left to right is: SWPM result, ground truth, two constant KM theory. All SWPM results are part of the untrained data. The delta values on top of our results and that of the two constant KM theory indicate the color distance to ground truth, the smaller this value is mean the more accurate the prediction is. Our prediction is more accurate in 12 out of all 15 examples shown in here.}
\label{fig:2_KM}       
\end{figure}

To evaluate our results and the results of the two-constant KM theory, we select 15 data untrained to predict our result and compute the result of the two-constant KM theory. In Fig. \ref{fig:2_KM}, our results, twelve cases out of fifteen, have smaller color distance value so being better than two-constant KM theory.

%
\FloatBarrier

%

\section{Applications}

For novice painters, color mixing is difficult because they must understand how to portray the colors of scenery or still life from a limited palette, which generally has between 12 to 18 pigments. Thus, we construct a system \cite{egp.20181008} called a Smart Palette to help them to acquire color mixing skills in Sect. \ref{SmartPalette}. Then, to prove our system, Smart Palette, can be applied to learn color mixing and the results of using the system to mix pigments is obviously better than a learning theory of traditional color mixing, Iteen's color wheel \cite{itten1970elements}, in Sect. \ref{UserStudy}.

\subsection{Smart Palette}
\label{SmartPalette}

\begin{figure}[htbp]
  \includegraphics[width = 1.0\textwidth]{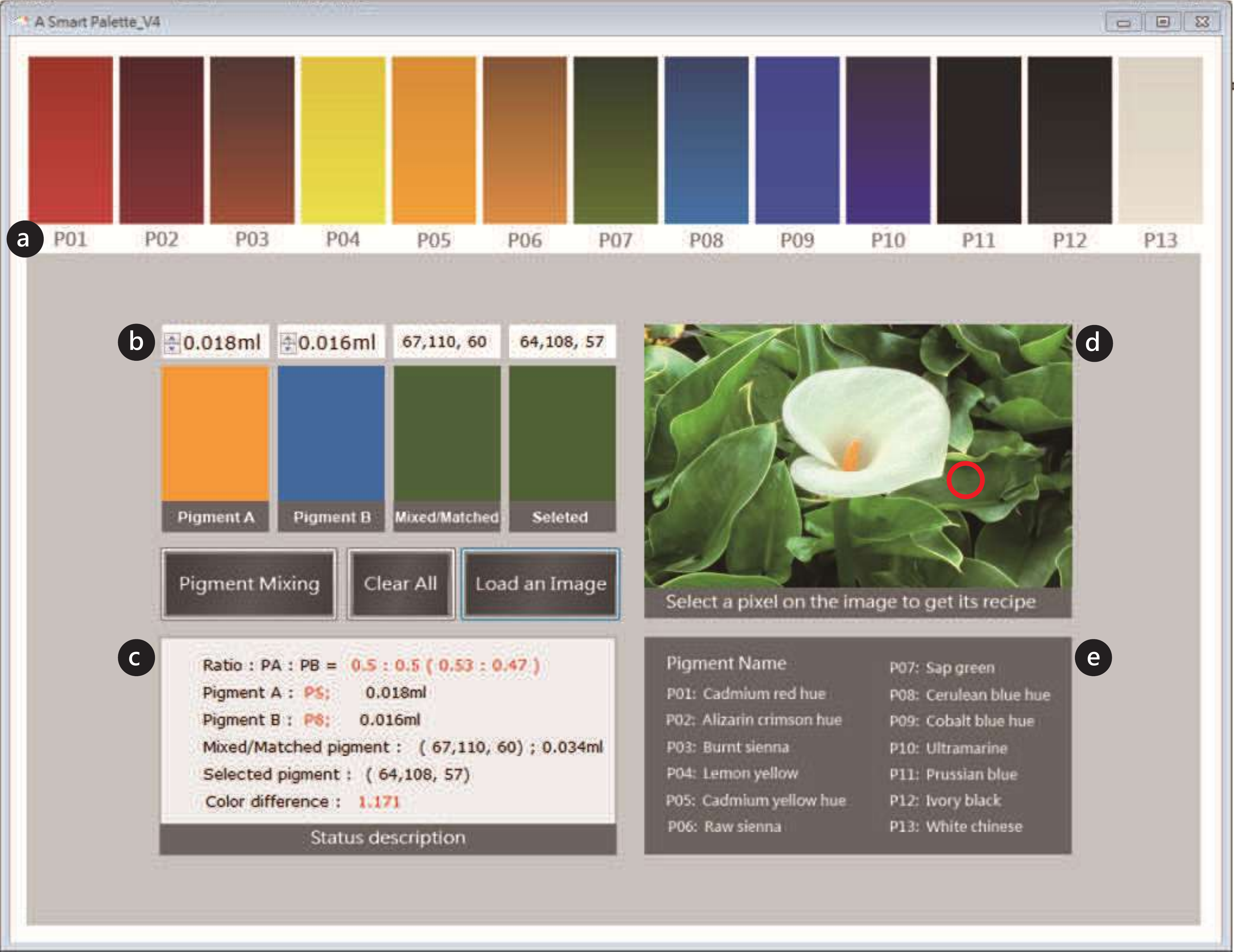}
\caption{Smart Palette. It is divided into five parts. In part (a), users can select 13 pigments with different quantities; each grid is a gradation, where at the top is the color of 0.16 mL per grid and at the bottom is the color of 0.01 mL per grid. Part (b) shows a result of color mixing or color matching and its ingredient pigment. The user can find more information about color mixing or color matching in part (c) and select a target color from an image in part (d). Finally, users can obtain the pigment name in part (e) and then use them to mix the color in the real world.}
\label{fig:Palette}       
\end{figure}

There are three steps in the construction for the Smart Palette: First, we interpolate the measured data of 13 primary pigments with 12 different quantities at 0.002 intervals. This step expands the 156 entries of the measured pigment data to 988 entries in total, which provides the system with more components to make predictions and thus increase both the quantity and quality of recipe our system can provide. Secondly, we prepare the input layer for the SWPM prediction model from the interpolated measured data. We use permutation on our interpolated measured data, which gives us a total of 976,144 sets (988 $\times$ 988) pigment mixtures-to-be-predicted, alongside and absolute reflectance of the white-backing paper; the procedure of predicting mixtures requires approximately 1 hour with an Intel(R) Xeon(R) E5-2650 v3 with a 2.30 GHz CPU. Finally, the prediction mixtures are transformed into Lab and RGB values to build a look-up table; thus, when a user picks a pixel from an image loaded in the Smart Palette, as shown in Fig \ref{fig:Palette} (d), the system will find a nearest RGB value from the look-up table by Lab distance as the matched color and return its corresponding recipe to the user. Recipes information display area can be seen as shown in Fig. \ref{fig:Palette} (b) and (c). 

The function mentioned above, which a user receives pigment mixture recipe for a desired target color by clicking, is one of the two functions in our Smart Palette system. Fig. \ref{fig:Palette} shows the interface of Smart Palette. A user can input an image into the system, perhaps one of scenery or still life which he/she wants to draw; then, he/she can select any pixel from the image; after that, he/she will obtain a recipe for the color of the chosen pixel, consisting two primary pigments and their corresponding quantities needed for the recreation of color with real life watercolor pigment. For example, a recipe is displayed in Fig. \ref{fig:Palette} (b) and (c): an RGB value (64, 108, 57) picked from within the red circle of the input image, as seen Fig. \ref{fig:Palette} (d), can be matched by another RGB value (67, 110, 60) in the color space of 13 watercolor pigments, and its ingredients are 0.018 mL of cadmium yellow hue, $\rho_5$, and 0.016 mL of cerulean blue hue, $\rho_8$. Searching time of the match color takes 1 second on a desktop computer (an Intel(R) Xeon(R) E3-1231 v3 with a 3.40 GHz CPU is used). 

For the current version of the Smart Palette, there are four recommendations points that MUST be implemented in order to obtain the most effective and accurate recipes and mixture results: 

\begin{enumerate}
\item The light source matters. The photograph used as the input image should be taken under a light source of daylight or overcast(a color temperature of 6500K). Because d65 is a light source of standard RGB monitor, we use d65 convert a prediction mixture represented a spectrum with 41 sampled nm into an RGB value. Thus, images photographed under such condition will generate the most correct recipe to help a user to mix pigments in the real world. 

\item When recreating a pigment mixture with recipe, focus on the primary pigment ratios, NOT the pigment volume (mL). To be clear, taking the pigment volume as instructed by the recipe DO allow the users to obtain correct mixtures; however, taking a correct volume down to 0.01mL or less requires tool, such as a syringe, to measure. Considering the actual application of color mixing, we recommend a simple eyeballing of the primary pigment ratios to mix pigments on a palette; once the mixture is complete, users can then decide how much of the mixture quantity to take to coat on a paper.

\item Find pixels with a recipe that has a smaller ratio range gap. To be specific, the pigment ratio gap between PA and PB must be less than 0.5 ($|PA - PB|$ < 0.5) to receive the best results. This is because Data type M. contains data of three mixing ratios (1:1, 1:2 and 2:1) currently, so our prediction model of color mixing can predict more accurate mixture between or close to those ratios. For example, the recipe shown in Fig. \ref{fig:Palette} (c) has a very small ratio gap, which means it is a good recipe. ( $|0.5 - 0.5|$ is less than 0.5)  

\item Find pixels with a recipe that has a smaller color difference. A small value of color difference means a high similarity between the selected pixel and its match. Using recipes with the smaller color difference level give the use more chance to recreate the color of the chosen pixel.

\end{enumerate}

\begin{figure}[htbp]
  \includegraphics[width = 1.0\textwidth]{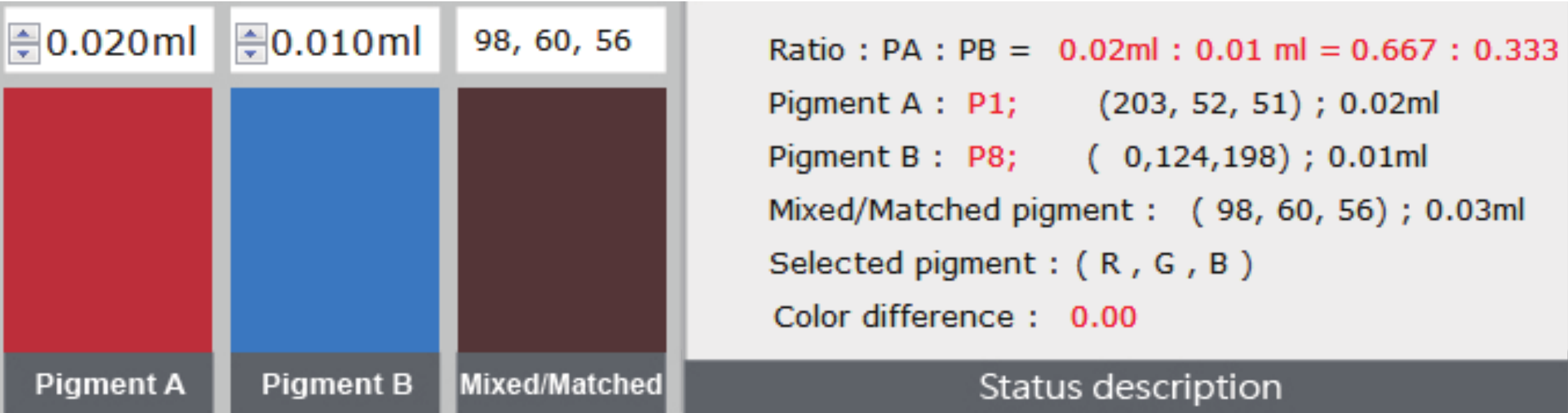}
\caption{A result of color mixing. When 0.02 mL per grid of $\rho_1$ and 0.01 mL per grid of $\rho_8$ are selected, our system displays their RGB values and the RGB value of their mixed color. }
\label{fig:Palette_mix}       
\end{figure}

The other function in our system is color mixing. In Fig. \ref{fig:Palette_mix}, users can select two pigments from Fig. \ref{fig:Palette}(a); then, the result of color mixing will be shown. In Sect. \ref{UserStudy}, we will do a user evaluation to understand  whether Smart Palette can help users mix pigments better than a traditional method.
 
%
%
%

\FloatBarrier
\subsection{User evaluation}
\label{UserStudy}
The Smart Palette system utilizes the advancements in hardware (low-cost spectrometer) and computation tool (accessible DNN tool) and creates an accurate and easy way to learn watercolor pigment mixing. This is not an overstatement but one validated by actual users. 

\subsubsection{Evaluation procedure}

\begin{figure}[htbp]
\centering
  \includegraphics[width = 1.0\textwidth]{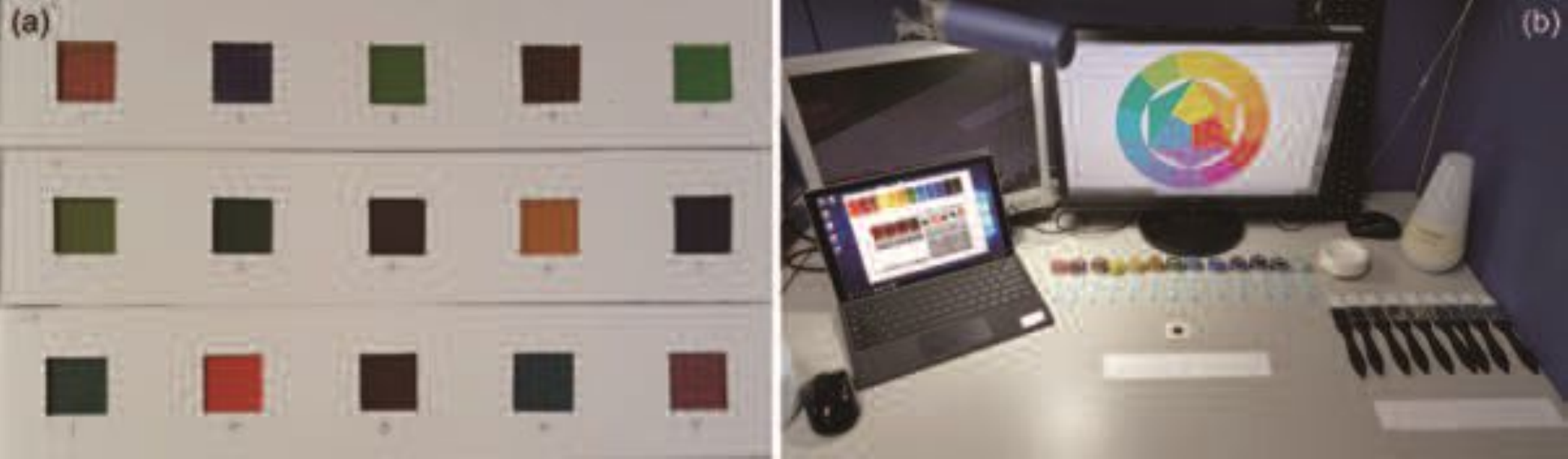}
\caption{The Target Color Card sample and evaluation environment setup. (a) the 3 Target Color Cards which contains a total of 15 target colors. The assignment of these Cards to evaluation users is permutated so each color is used equally for each Task. (b) the complete setup for evaluation, including the 13 primary pigments, pigment sampling scoop, disposable paper palette, silicon brush, Answer Cards,  image of Itten's color wheel for the task session B, and the laptop loaded with our Smart Palette for task session C. }
\label{fig:UserPre}       
\end{figure}

We recruit 18 non-painter users between 20 to 35 years of age to participate in a user evaluation designed to compare our system to traditional method on the effectiveness of learning watercolor mixing. Users were asked to perform 3 sets of color mixing tasks, A, B, and C. Each task shares the same goal, which the user needs to choose 2 from the 13 provided pigments to create mixtures match closest to the target colors. There are a total of 15 target colors divided into 3 Target Color Cards, which contains 5 target colors on each card (see Fig. \ref{fig:UserPre} (a)).  These cards are assigned to the users and permutated to ensure each target color is used equally in each task.

Task A is where users perform color mixing entirely by intuition; Task B is performed after the user receives an introduction on Itten's color wheel theory; Task C is where users use our Smart Platte. Users perform these three tasks in the order of A, B, and then C.  

The performing of tasks requires the participant to first pick 2, and only 2, pigments for the attempt on recreating each target color. Once the 2 pigments are chosen, the user proceeds on the actual pigment mixing by sampling the chosen pigments with scoop and adjusting the mixing ratio on disposal paper palettes. The mixture pigment is then applied onto one of the five 1.5 $cm^{2}$ grid on an Answer Card after the user is done mixing, either from reaching satisfaction or otherwise. 

\begin{figure}[htbp]
\centering
  \includegraphics[width = 1.0\textwidth]{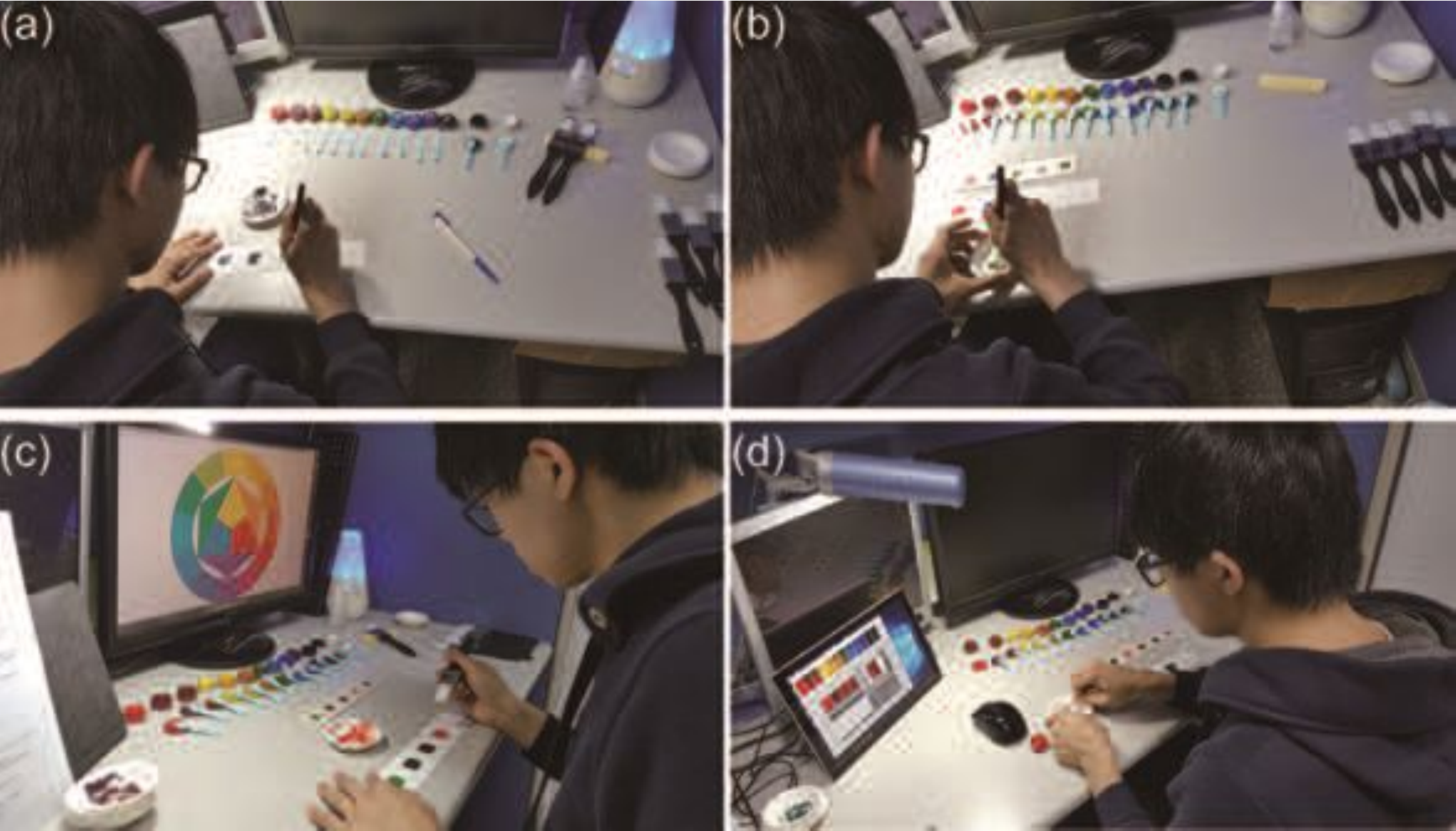}
\caption{Actual images of evaluation process. (a) the Test Run, where participants familiarize themselves with evaluation procedure and tools. The amount of time spent in this session ranged from 6~10 minutes. (b) Task A: Mixing pigment by intuition. (c) Task B: Mixing after studying Itten's color theory (d)Task C: Mixing according to the recipes provided by the Smart Palette, which participants operates the system themselves after learning the interface.}
\label{fig:UserStudy}       
\end{figure}

Before the actual evaluation, there is a test run to familiarize participants with the whole procedure, from observing the target color from a Target Color Card, choosing the 2 pigments for the mixture, sampling and adjusting mixing ratio of the pigments, and lastly, coating of the mixture pigment onto the Answer Card. Fig. \ref{fig:UserPre} demonstrates the actual Target Color Card and the environment setup, including all tools provided to the evaluation participants. Images in Fig. \ref{fig:UserStudy} are photographed during one of the actual evaluation, shows all task procedures. 

%

%

\subsubsection{Evaluation result}

Every participant, after completing 5 mixtures each for all 3 tasks, ends up producing 15 pigment mixtures in total. The amount of time spent on completing all 3 tasks is 71 minutes on average, excluding the task introduction, Q$\&$As prior to the evaluation, and the test run.  

These mixtures are then measured for accuracy comparisons against their corresponding target color in Lab distance; the greater the distance is signified the more user's mixture is dissimilar from the correct answer (target color), whereas the smaller distance means the better the user performs in color mixing.  All Lab distance scores from all 18 evaluation participants are sorted and shown in Table \ref{3Tasks}. In table \ref{3Tasks}, each user has 3 Lab distance scores and each score is averaged from all 5 answers in one task. The scores of individual users on each task is again calculated for the total task average at the bottom row of the table. The total task score reads: Task A: 13.97; Task B: 14.39; Task C: 7.86. From the figures along, we can find that there are general improvements in Task C, where users have access to our Smart Palette tool in color mixing.

In order to determine whether the improvement of performance is statistically significant, the 3 total task scores (simplified as A, B, C) are taken 2 at a time to examine with two-tails Student's t-tests. In other words, we run a total of 3 t-tests and their null hypotheses, alternative hypotheses, and the t-values as follows: 

\begin{enumerate}
\item Null hypothesis: A = B; alternative hypothesis: A $\neq$ B, t-value = 0.2491
\item Null hypothesis: B = C; alternative hypothesis: B $\neq$ C, t-value = 4.5265
\item Null hypothesis: C = A; alternative hypothesis: C $\neq$ A, t-value = 4.8780
\end{enumerate}

Through the results of these three two-tails t-tests, we can conclude with great confidence (in fact, with a confidence level greater than 99$\%$) that the improvement of user's accuracy in recreating pigment mixture that matches a targeted color is very significant. That is to say, non-painter users do perform much more accurately at pigment mixing when assisted by our Smart Palette, as compared to that of using intuition or with Itten's color wheel.

\begin{table}[htbp]
\centering
\caption{The task scores of all 3 Tasks (5 mixture per task) from all 18 evaluation participants, shown both as individuals average (per participant, per task) and as the overall average (all participant, per task). It can be observed that the difference of accuracy (Lab distance) is very marginal between Task A and Task B. On the other hand, the difference of Task A, B to Task C is great as 6.53.}
\vspace{0.5cm}
\label{3Tasks}
\begin{tabular}{|c|c|c|c|}
\hline
                                & \cellcolor[HTML]{C0C0C0}Task A & \cellcolor[HTML]{C0C0C0}Task B & \cellcolor[HTML]{C0C0C0}Task C \\ \hline
User 01                         & 12.902                         & 12.63                          & 3.802                          \\ \hline
User 02                         & 18.264                         & 11.996                         & 6.476                          \\ \hline
User 03                         & 17.152                         & 10.712                         & 5.206                          \\ \hline
User 04                         & 21.16                          & 12.688                         & 10.268                         \\ \hline
User 05                         & 21.958                         & 15.722                         & 6.736                          \\ \hline
User 06                         & 20.6                           & 13.038                         & 4.368                          \\ \hline
User 07                         & 12.864                         & 11.25                          & 6.858                          \\ \hline
User 08                         & 11.076                         & 12.716                         & 10.51                          \\ \hline
User 09                         & 10.34                          & 22.418                         & 9.006                          \\ \hline
User 10                         & 13.71                          & 7.576                          & 15.112                         \\ \hline
User 11                         & 13.422                         & 15.242                         & 12.938                         \\ \hline
User 12                         & 14.114                         & 8.986                          & 12.006                         \\ \hline
User 13                         & 9.786                          & 14.104                         & 5.678                          \\ \hline
User 14                         & 8.202                          & 20.144                         & 5.192                          \\ \hline
User 15                         & 13.372                         & 14.952                         & 6.336                          \\ \hline
User 16                         & 11.42                          & 13.632                         & 7.87                           \\ \hline
User 17                         & 12.064                         & 23.72                          & 5.676                          \\ \hline
User 18                         & 9.12                           & 17.418                         & 7.366                          \\ \hline
\rowcolor[HTML]{DAE8FC} 
\cellcolor[HTML]{FFFFFF}Average & 13.97                          & 14.39                          & 7.86                           \\ \hline
\end{tabular}
\end{table}
\section{Conclusion and future works}

\subsection{Conclusion}
\label{Conclusion}
In this paper, we established a dataset on physical watercolor pigments. NTU WPSM dataset contains a total of 1,222 labeled data, produced from careful sample making and measurement, which is the base of SWPM prediction model and our Smart Palette tool system and still have potential to develop further upon. In our experiment, 83$\%$ of the data in our test scores less than 5 $\Delta E^{*}_{ab}$, which is below the threshold for being determined as a mismatch. In a comparison experiment, most of the prediction mixtures of our prediction model are closer to ground truth than the color mixing results of the two-constants KM theory. In addition, in order to examine the effectiveness of the Smart Palette system, we design a user evaluation which untrained users perform pigment mixing with three methods: by intuition, based on Itten's color wheel, and with access to the Smart Palette. Assessing by the results of color distance scores and the t-values from t-tests, it is clear that the user produce pigment mixtures that are significantly closer to target colors when following the Smart Palette recipe. Our system, the Smart Palette, demonstrates effectiveness in helping users to learn and perform better at color mixing than the aids traditional methods could provide.  

Of course, our work is not without limitations at its current state. The accuracy we pride ourselves with only holds the most truth when the target color, regardless in a mixing or matching scenario, is within the color space based on the 13 Winsor and Newton pigments we selected for this study. Furthermore, mixtures or matches that require more than 2 of the selected 13 primary pigments to achieve also loses precision under the current model. Another issue in our work is the symmetry issue, which is intrinsic in DNN. We do, however, recognize these limitations as somewhere excusable, since the former limitation can be surpassed with just more data and time because the model and procedure is proven viable; and the latter issue can be amended in application because, even for cases with the greatest asymmetry, one of the predictions still matches the ground truth closely.

Furthermore, we believe that the value of this study surpass the study, such as the data set, the prediction model, or even the Smart Palette tool, itself.  It present one viable research procedure that could enable other academics to attempt studies in the field of color and spectrum; it also present an approach that incorporate the physics of seeing color into the field of Computer Graphics, such as color mixture prediction and color adjustment. Elaboration on these two points is as follows:

\begin{enumerate}
\item Utilization of DNN for expanding costly spectrum data: In our work, DNN is employed to expand 1,222 extremely labor-heavy data on semitransparent watercolor pigment into a massive 976,144 dataset. The procedure breaks down to first creating and measuring partial spectrum data in key ratio; then takes 80$\%$ of the measured data for training and 20$\%$ for testing to complete the prediction model; lastly, the prediction model is ready to make prediction for a large amount of spectrum data that were not directly measured yet proved to be accurate enough to test out different concept and research hypotheses. 

\item An interdisciplinary approach which grants more flexibility in later-adjustment by denoting colors in intrinsic physical property:  The reason why we painstakingly develop our prediction model based on data taken directly from real pigments with spectrometer, instead of digitally generated color blocks, is because it allows us to accommodate more external factors that exist between the object to our eyes. The physics of seeing color involves the object, light source, and the observer. Our approach which denoting the colors in spectrum isolates the data to the object itself, which means the prediction results can be adjusted easily with different light source and observer factors.
\end{enumerate}

To sum up, our Smart Palette tool has enough accuracy and practicality to be considered by watercolor learners as a new option in learning color mixing. More importantly, it shows that spectrum data, something traditionally takes enormous time and labor to acquire, can use DNN to receive good-enough predicted data that allows concepts and ideas to be tested. In addition, our approach reveals the advantage in adjustment flexibility by denoting colors in more essential form. This interdisciplinary approach between physics and computer science hopefully can inspire more cross-field studies in the future.

\subsection{Future works}
\label{Future}
For future works, some can be developed with our existing data directly and some more can be explored by applying similar procedures and approaches as described in Sect. \ref{Conclusion}.
 
For direct developments, other researchers are welcome to construct prediction models with even better accuracy with our open dataset. Our accurate and natural-looking watercolor pigment prediction results can also be used in watercolor rendering to provide more realistic rendering results.
In terms of migrating our procedure, the way we measure and construct prediction model and create workable color mixing tool is reproducible for different coloring media, such as acrylic paint, ink, and oil paint, etc.

\renewcommand\arraystretch{1.6}
\begin{table}
\caption{Hypothetical future research projects that could apply our research procedure and interdisciplinary approach.}
\centering
\label{5case}
\resizebox{\textwidth}{!}{%
\begin{tabular}{|l|c|c|c|c|}
\hline
case & Color property of Object & Environmental Factors & Observer Perception & \multirow{6}{*}{\begin{tabular}[c]{@{}c@{}}How Colors perceived\\  by observer(s)\end{tabular}} \\ \cline{1-4}
A & \begin{tabular}[c]{@{}c@{}}Reflectance and transmittance \\ of media (watercolor)\end{tabular} & \begin{tabular}[c]{@{}c@{}}Light source \\ color temperature (D65)\end{tabular} & \begin{tabular}[c]{@{}c@{}}Color recognition \\ for standard observer  (CMF)\end{tabular} &  \\ \cline{1-4}
B & RGB value & \begin{tabular}[c]{@{}c@{}}Panel and software filter \\ of mobile device\end{tabular} & \begin{tabular}[c]{@{}c@{}}Color recognition \\ for standard observer\end{tabular} &  \\ \cline{1-4}
C & RGB value & \begin{tabular}[c]{@{}c@{}}Panel and software filter\\  of mobile device\end{tabular} & \begin{tabular}[c]{@{}c@{}}Color recognition \\ for aging observer\end{tabular} &  \\ \cline{1-4}
D & RGB value & \begin{tabular}[c]{@{}c@{}}Panel and software filter\\  of television\end{tabular} & \begin{tabular}[c]{@{}c@{}}Color recognition \\ for aging observer\end{tabular} &  \\ \cline{1-4}
E & RGB value & \begin{tabular}[c]{@{}c@{}}Panel and software filter \\ of television\end{tabular} & \begin{tabular}[c]{@{}c@{}}Color recognition \\ for color-blind\\   observer\end{tabular} &  \\ \hline
\end{tabular}%
}
\end{table}

The part with the most potential is perhaps our interdisciplinary approach. We borrow the physic of seeing color, which roughly consist the object, light source, and observer, to fit application in Computer Graphics. Table \ref{5case} demonstrates how we analyze how colors are seen, alongside many of few cases of future works possible: 

Case A in Table \ref{5case} describes the works of this very study. The colors that are finally perceived by an observer can be broken down three as components: reflectance and transmittance of pigments, observer perception representing by the color matching function (CMF), and light source of D65. Within these three components the light source being the most easily swappable, which can be adjusted to D55 or others to simulate how the same pigment will be seen under different observation environments.  

Consider a scenario with Case B, where two different brands of cellphones (PhoneI $\&$ PhoneA) are input with the same RGB, and the observer perception condition assumed to be standard. Under such circumstance, the factor left determining how observer sees the color on these two cellphones is at their screen panel and software filter. Theoretically, with the observer condition remains constant, the colors on two brands of mobile devices should be perceived as the same if adjustments can be made so that the rest two components (roughly simplified as "Panel(f(RGB)") from PhoneI and PhoneA become identical. In practice, researchers can measure the screen directly with spectrometer to receive data on "Panel(f(RGB)". With the RGB value being manipulatable and the observer factor being a constant, a function that could match the "Panel(f(RGB)" on PhoneA to PhoneI, or vice versa, should be discoverable. This can be used to change how colors are seen by users from one device to another to match the user's preference. The same concept can be applied to Case C, D, and E, of different type of devices or different perception conditions.    

Furthermore, assuming that a user prefers how one image looks on an iPhone over an Android device, because the color of one image may feel more aesthetically pleasing. By going through the procedures: 

\begin{enumerate}
\item Measure and construct a dataset on two phones RGB values, Panel and software filter; 
\item Expand data with prediction model (low-cost method to receive data that were not directly measured) and ; 
\item Find a function that change the value of Android's "Panel(f(RGB)" to that of the iPhones' by recalibrating the RGB values by pixels for the whole image. 
\end{enumerate}

\begin{figure}[htbp]
\centering
  \includegraphics[width = 1.0\textwidth]{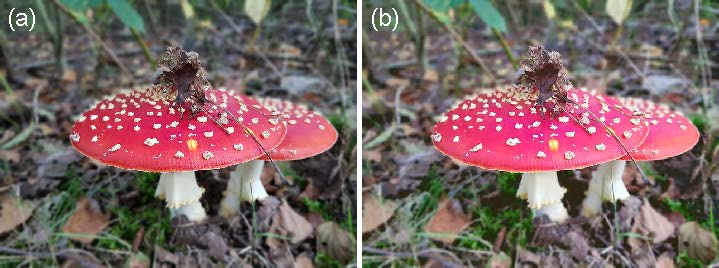}
\caption{A comparison between the original image  and the pixel-adjusted image for Android to look like iPhone. (a) the original image, which is from a website, pixabay, and its copyright is free for commercial use and no attribution required. (b) the pixel-adjusted image of the original. }
\label{fig:FW}       
\end{figure}

Through this procedure, the image should be able to look as similarly pleasing to the user on an Android phone as if it is on an iPhone. Fig. \ref{fig:FW} (a) shows an example image, in which after being adjusted according to the "Panel(f(RGB)" of an iPhone, the image output is shown in Fig. \ref{fig:FW} (b). The evaluation needs a final step, which is to put Fig. \ref{fig:FW} (a) to an iPhone while putting Fig. \ref{fig:FW} (b) to an Android and compare the image on two devices side-by-side.  The study described above is actually in progress currently. 

%

\bibliographystyle{unsrt}  
 \bibliography{references}  






\end{document}